\documentclass[12pt]{iopart}
\usepackage{iopams}  
\expandafter\let\csname equation*\endcsname=\relax
\expandafter\let\csname endequation*\endcsname=\relax
\usepackage{amsmath,amssymb,amsthm}
\usepackage{mathrsfs}

\usepackage{amsfonts}
\usepackage{epstopdf}
\usepackage{graphicx}
\usepackage{url}
\usepackage{dsfont}
\usepackage{xcolor}
\usepackage{subcaption}
\usepackage{setspace}
\usepackage{booktabs}
\usepackage{tablefootnote} 
\usepackage{lineno}
\usepackage{soul}

\let\<=\langle
\let\>=\rangle
\def\~#1{{\mbox{#1}}}

\let\-=\mathbf

\def\circ{\ifmmode\mathchar"220E\else$\mathchar"220E$\fi}
\def\@#1{{\mathcal #1}}
\usepackage[inline]{enumitem}
\usepackage{threeparttable}
\usepackage{diagbox}
\usepackage{lipsum}
\usepackage{amsfonts}
\usepackage{graphicx}
\usepackage{multirow}
\usepackage{epstopdf}
\usepackage{amsopn}

\usepackage{algorithm}
\usepackage{algpseudocode,float}
\ifpdf
  \DeclareGraphicsExtensions{.eps,.pdf,.png,.jpg}
\else
  \DeclareGraphicsExtensions{.eps}
\fi

\usepackage{enumitem}
\setlist[enumerate]{leftmargin=.5in}
\setlist[itemize]{leftmargin=.5in}

\begin{document}
\title[]{Deep Unrolling Networks with Recurrent Momentum Acceleration for Nonlinear Inverse Problems}
\author{Qingping Zhou$^1$, Jiayu Qian$^1$, Junqi Tang$^2$, Jinglai Li$^{2,*}$}
\address{$^1$School of Mathematics and Statistics, HNP-LAMA, Central South University, Changsha 410083, China\\
$^2$School of Mathematics, University of Birmingham, Edgbaston, Birmingham B15 2TT, UK}
\ead{j.li.10@bham.ac.uk}
\vspace{10pt}

\begin{abstract}
Combining the strengths of model-based iterative algorithms and data-driven deep learning solutions, deep unrolling networks (DuNets) have become a popular tool to solve inverse imaging problems. Although DuNets have been successfully applied to many linear inverse problems, their performance tends to be impaired by nonlinear problems. Inspired by momentum acceleration techniques that are often used in optimization algorithms, we propose a recurrent momentum acceleration (RMA) framework that uses a long short-term memory recurrent neural network (LSTM-RNN) to simulate the momentum acceleration process. The RMA module leverages the ability of the LSTM-RNN to learn and retain knowledge from the previous gradients. We apply RMA to two popular DuNets -- the learned proximal gradient descent (LPGD) and the learned primal-dual (LPD) methods, resulting in LPGD-RMA and LPD-RMA, respectively. We provide experimental results on two nonlinear inverse problems: a nonlinear deconvolution problem, and an electrical impedance tomography problem with limited boundary measurements. In the first experiment we have observed that the improvement due to RMA largely increases with respect to the nonlinearity of the problem. The results of the second example further demonstrate that the RMA schemes can significantly improve the performance of DuNets in strongly ill-posed problems. 
\end{abstract}

\vspace{2em}

\maketitle

\newpage
\section{Introduction}\label{sec:intro}
Many image processing tasks can be cast as an inverse problem, i.e., to recover an unknown image  $x$ from indirect measurements $y$
\begin{equation}\label{eq:forward}
y=\mathcal{A} (x)+\epsilon, 
\end{equation}
where $x\in X$, $y\in Y$, $\mathcal{A}$ represents a forward measurement operator and $\epsilon$ is the observation noise. Problems that can be formulated with Eq.~\eqref{eq:forward} include denoising~\cite{houdard2018high}, compressive sensing~\cite{zhang2018ista}, computed tomography reconstruction~\cite{adler2018learned,baguer2020computed}, phase retrieval~\cite{shechtman2015phase}, optical diffraction tomography~\cite{sung2009optical}, 
electrical impedance tomography~\cite{seo2019learning,colibazzi2022learning,wang2024comparative} and so on.

A common challenge in solving the inverse problems is that they are typically ill-posed and 
as such regularization techniques are often used to ensure a unique and stable solution. Classical regularization techniques introduce an explicit regularization term in the formulation, yielding a variational regularization problem that can be solved with iterative algorithms, such as the alternating direction method of multipliers (ADMM)~\cite{boyd2011distributed} and the primal-dual hybrid gradient (PDHG) method~\cite{chambolle2011first}. Although such methods can obtain well-posed solutions of the inverse problems, they have a number of limitations: most notably, inaccuracy regularizing assumption, need for parameter tuning, and mathematical inflexibility. A possible remedy to these limitations is to implicitly regularize the inverse problem by replacing certain modules in the iterative procedure with deep neural networks~\cite{adler2017solving,adler2018learned,zhang2018ista,xu2024enhancing}, a type of method commonly referred to as the deep unrolling networks (DuNets).

DuNets, pioneered by Gregor and LeCun in~\cite{gregor2010learning}, have achieved great empirical success in the field of inverse problems and image processing~\cite{zhang2019dynamically,monga2021algorithm,tang2022accelerating} in the past few years. 
DuNets combine traditional model-based optimization algorithms with learning-based deep neural networks, yielding an interpretable and efficient deep learning framework for solving inverse imaging problems. It should be noted that most of the aforementioned studies have focused on solving inverse problems with linear or linearized forward operators. On the other hand, there has been relatively little research on applying DuNets to nonlinear inverse problems, such as optical diffraction tomography and electrical impedance tomography (EIT). This paper attempts to bridge this gap by applying DuNets to the nonlinear inverse problems.
In such problems, nonlinearity may pose additional difficulty for the DuNet methods, as the gradient of the forward operator varies significantly as the iterations proceed. Intuitively, the performance of DuNets may be improved if the previous gradient information is included. In this work we will draw on momentum acceleration (MA), an acceleration strategy commonly used in optimization algorithms, and the recurrent neural networks (RNN) techniques to improve the performance of the deep unrolling networks. Specifically, we propose a recurrent momentum acceleration (RMA) module that utilizes a long short-term memory recurrent neural network (LSTM-RNN) to represent the momentum acceleration term. The RMA module exploits the LSTM-RNN's capacity to remember previous inputs over extended periods and learn from them, thereby providing information from the entire gradient history. 
In particular we apply it to two popular DuNets: the learned proximal gradient descent (LPGD)~\cite{mardani2018neural}  and the learned primal-dual (LPD)~\cite{adler2018learned}. We refer to the resulting algorithms as LPGD-RMA and LPD-RMA respectively. 
It should be noted that several existing works propose to use the iteration history to improve the performance of the unrolling algorithms. For example,  \cite{adler2018learned} extends the state space to allow the algorithm some ``memory'' between the iterations, and \cite{hosseini2020dense} develops a history-cognizant unrolling of the proximal gradient descent where the outputs of all the previous regularization units are used. As a comparison, our RMA method employs the previous gradients that are combined via a flexible RNN model learned from data. 

The remainder of this paper is organized as follows. In section~\ref{sec:DuNets}, we review two widely used types of deep unrolling models: the LPGD and the LPD methods. In section~\ref{sec:DuNets-RMA}, we present our RMA formulation, and incorporate it with both LPGD and LPD, yielding LPGD-RMA and LPD-RMA. Numerical experiments performed on two nonlinear inverse problems are reported in section~\ref{sec:exp}. Finally section~\ref{sec:conclusion} offers some concluding remarks.

\section{Deep unrolling networks}\label{sec:DuNets}
We start with the variational formulation for solving inverse problems of the form~\eqref{eq:forward}. 
These methods seek to solve the following  minimization problem 
that includes a data consistency term $\mathcal{D}(\cdot,\cdot): Y \times Y \rightarrow \mathbb{R}$ 
and regularization term $\@R(\cdot): X \rightarrow \mathbb{R}$:
\begin{equation}\label{eq:optim}
\underset{x \in X}{\arg \min }~
\mathcal{D}(\mathcal{A}(x), y)+\lambda \mathcal{R}(x),
\end{equation}
where $\lambda$ is the regularization parameter balancing $\mathcal{R}$ against $\mathcal{D}$.
The regularizer $\mathcal{R}$  encodes prior information on $x$ representing desired solution properties. Common regularization functions include Tikhonov regularization, total variation (TV), wavelets, and sparsity promoting dictionary~\cite{benning2018modern}, to name a few.  Besides Tikhonov, all the aforementioned regularization functions are non-smooth, and as a result Eq.~\eqref{eq:optim} is typically solved by the first-order algorithms, such as the proximal gradient descent~(PGD) algorithm~\cite{combettes2011proximal}, the variable splitting scheme~\cite{boyd2011distributed} and the primal-dual hybrid gradient (PDHG) method~\cite{chambolle2011first}. 
These algorithms typically involve rather expensive and complex iterations. 
The basic idea of DuNets is to use a learned operator, represented by a deep neural network, to model the iterations. We here focus on two archetypes of the deep unrolling models: the LPGD method with both shared \cite{mardani2018neural,lohit2019unrolled} and independent weights~\cite{gupta2018cnn,cherkaoui2020learning}, and the LPD method~\cite{adler2018learned}. 

\subsection{Learned proximal gradient descent method}
Starting from an initial value $x_0$, PGD performs the following iterates until convergence:
\begin{subequations}\label{eq:pgd}
\begin{align}
s_{t} &= x_{t-1}-\alpha_t \nabla_{x_{t-1}} \mathcal{D}(\mathcal{A}(x_{t-1}), y), \label{eq:gd} \\ 
x_{t} &= \mathcal{P}_{\lambda \@R}\left(s_{t} \right), \label{eq:proximal}
\end{align}
\end{subequations}
where $\alpha_t$ is the step size  and the proximal gradient descent $\mathcal{P}_{\lambda \@R}(\cdot)$ is defined  by
\begin{equation*}
\@P_{\lambda \@R}(x)=\underset{x^{\prime} \in X}{\arg \min }~\frac{1}{2}\left\|x^{\prime}-x\right\|_{X}^{2}+\lambda \mathcal{R} \left(x^{\prime}\right).
\end{equation*}

Simply speaking, the LPGD method replaces the proximal operators $\@P_{\lambda \@R}$ with a neural network $\Psi_{\theta_t}$~\cite{gupta2018cnn,cherkaoui2020learning,yang2022dynamic},  and thus it allows one to learn how to effectively combine the previous update with the gradient update direction instead of a pre-determined updating scheme. 
The complete algorithm of LPGD is outlined in Algorithm~\ref{alg:lpgd}.
In the standard LPGD algorithm, the network in each iteration uses its own weights $\theta_t$,
and a popular variant of the method is to use shared weights over all the networks, {i.e.}, restricting $\theta_1=...=\theta_T$.
We refer to~\cite{mardani2018neural,lohit2019unrolled} for more details.

\begin{algorithm}[H]
      \small
      \caption{LPGD algorithm}
      \label{alg:lpgd}
        \begin{algorithmic}[1]
           \item[] \textbf{Input:} $x_0 \in X$
           \item[] \textbf{Output:} $x_T$  
           \For{$t=1,\ldots,T$}
             \State{$g_{t-1} = \nabla_{x_{t-1}} \mathcal{D}(\mathcal{A}(x_{t-1}), y)$}
             \State{$x_t = \Psi_{\theta_t}(x_{t-1}, g_{t-1})$}
           \EndFor
        \end{algorithmic}
\end{algorithm}

\subsection{Learned primal-dual method}\label{sec:lpd} 
Adler and $\ddot{\mathrm{O}}$ktem first introduced the partially learned primal-dual approach as an extension of iterative deep neural networks in~\cite{adler2017solving} and further elaborated it into the LPD approach in~\cite{adler2018learned}. 
The method is based on PDHG, another popular algorithm for solving the non-differentiable optimization problem \eqref{eq:optim}.
The PDHG iteration is given by
\begin{equation}\label{eq:PDHG}
\left\{\begin{array}{l}
u_{t+1} =\operatorname{prox}_{\sigma \mathcal{D}^*}\left(u_t+\sigma \mathcal{A}\left(\bar{x}_t\right)\right) \\
x_{t+1} =\operatorname{prox}_{\tau \mathcal{R}}\left(x_t-\tau\left[\partial \mathcal{A}\left(x_t\right)\right]^*\left(u_{t+1}\right)\right) \\
\bar{x}_{t+1} =x_{t+1}+\gamma\left(x_{t+1}-x_t\right)
,\end{array} 
\right.
\end{equation}
where $\sigma, \tau, \gamma$ are predefined parameters, $\mathcal{D}^*$ denotes the Fenchel conjugate of $\mathcal{D}$,  and $\left[\partial \mathcal{A}\left(x_t\right)\right]^*$ is the adjoint of the Fr\'{e}chet derivative of $\mathcal{A}$ at point $x_t$. 
The LPD method is built upon Eq.~\eqref{eq:PDHG},
and  the main idea is to replace $\operatorname{prox}_{\sigma \mathcal{D}^*}$ and $\operatorname{prox}_{\tau \mathcal{R}}$ with  neural network models that are learned from data. 
Once the models are learned, the reconstruction proceeds via the following iterations: 
\begin{equation}\label{eq:LPD}
\left\{\begin{array}{l}
u_{t} =\Gamma_{\theta_{t}^{d}}\left(u_{t-1}, \mathcal{A}(x_{t-1}^{}), y\right) \\ x_{t} =\Lambda_{\theta_{t}^{p}}\left(x_{t-1},\left[\partial \mathcal{A}(x_{t-1})\right]^{*}\left(u_{t}^{}\right)\right)
,\end{array} \quad \text {for}~~~ t=1, \ldots, T.\right.
\end{equation}
The complete LPD algorithm is given in Algorithm~\ref{alg:lpd}.
It is important to note that the LPD method often enlarges both the primal and dual spaces to allow some ``memory'' between iterations~\cite{adler2018learned}. In particular, it defines  $x_t=[x_t^{(1)}, x_t^{(2)}, \ldots, x_t^{(N_{\text{primal}})}] \in X^{N_{\text{primal}}}$ and $u_t=[u_t^{(1)}, u_t^{(2)}, \ldots, u_t^{\left(N_{\text{dual}}\right)}] \in Y^{N_{\text{dual}}}$.
Additionally, $\Lambda_{\theta_{t}^{p}}: X^{N_{\text{primal}}} \times X^{N_{\text{primal}}} \rightarrow X^{N_{\text{primal}}}$ corresponds to dual and primal networks, which have different learned parameters but with the same architecture for each iteration.
A typical initialization is $x_{0}=[0,\cdots,0]$ and $u_{0}=[0,\cdots,0]$, where 0 is the zero element in the primal or dual space. We refer to~\cite{adler2018learned} for details on the LPD method.
    \begin{algorithm}[H]
            \small
           \caption{LPD algorithm}
           \label{alg:lpd}
            \begin{algorithmic}[1]
               \item[] \textbf{Input:} $x_0 \in X^{N_{\text{primal}}}, u_0 \in Y^{N_{\text{dual}}}$  
               \item[] \textbf{Output:} $x_T^{(1)}$  
               
               \For {$t=1,\ldots,T$}
                    \State $u_{t} =\Gamma_{\theta_{t}^{d}}\left(u_{t-1}, \mathcal{A}(x_{t-1}^{(2)}), y\right)$
                    \State $g_t = \left[\partial \mathcal{A}(x_{t-1}^{(1)})\right]^{*} (u_{t}^{(1)})$
                    \State $x_{t} =\Lambda_{\theta_{t}^{p}}\left(x_{t-1}, g_t\right)$
              \EndFor
            \end{algorithmic}
    \end{algorithm}
    
\section{Deep unrolling networks with  momentum acceleration}\label{sec:DuNets-RMA}
The conventional DuNets, exemplified by LPGD and LPD, only use the current gradient, ignoring a large amount of historical gradient data. As has been discussed, these methods can be improved by adopting the momentum acceleration (MA) strategies that are frequently used in optimization methods.
In this section we will present such momentum accelerated DuNet methods. 

\subsection{Momentum acceleration methods}\label{sec:momentum}
We here discuss the conventional explicit MA scheme and the one based on RNN. 

\subsubsection{Explicit momentum acceleration}
Momentum-based acceleration methods, like Nesterov’s accelerated gradient~\cite{nesterov1983method} and adaptive moment estimation~\cite{adam2015}, are well-established algorithms for speeding up the optimization procedure and have vast applications in machine learning~\cite{sutskever2013importance}. 
The classical gradient descent with MA utilizes the previous ``velocity'' $v_{t-1}$ at each iteration to perform extrapolation and generates the new update as:
\begin{subequations}\label{eq:momentum}
\begin{align}
v_t &= \gamma v_{t-1} - \eta 
 g_{t-1} \label{eq:vtma}\\
x_t &= x_{t-1} + v_t, 
\end{align}
\end{subequations}
where $g_{t-1}$ is the gradient of the objective function evaluated at $x_{t-1}$, $\eta$ is the step size,
and $\gamma \in[0,1)$ is the momentum coefficient controlling the relative contribution of the current gradient and the previous velocity.  
Eq~\eqref{eq:momentum} can be rewritten as:
\begin{equation}\label{eq:vt}
\begin{aligned}
    v_t &=\gamma v_{t-1} -\eta g_{t-1} 
    = \gamma (\gamma v_{t-2}-\eta g_{t-2} -\eta g_{t-1})\\
    &= \ldots 
    =\gamma^t v_0 
    -\gamma^{t-1}\eta g_0
    -\ldots
    -\eta g_{t-1},
\end{aligned}
\end{equation}
which shows that the current velocity is essentially a weighted average of all the gradients (assuming $v_0=0$). 
As mentioned above, the momentum coefficient $\gamma$ controls how much information from previous iterations is used to compute the new velocity $v_t$, and therefore needs to be chosen carefully for a good performance of the method. 
However, the optimal value for the parameter is problem-specific and typically requires manual tuning~\cite{sutskever2013importance,liu2020improved}.
 
\subsubsection{Momentum acceleration via RNN}
The conventional momentum method utilizes a fixed formula (a linear combination of all the gradients) to calculate the present velocity $v_t$.
In this section we introduce a more flexible scheme that uses the recurrent neural networks to learn the velocity term~\cite{hochreiter1997long}, which we refer to as the RMA method. 
In particular we use the LSTM based RNN, which is briefly described as follows. 
At each time step $t$ we employ a neural network to compute the ``velocity'' $v_t$.
The neural network has three inputs and three outputs. 
These inputs include the current gradient input $g_{t-1}$, the cell-state $c_{t-1}$ (carrying long-memory information) and the hidden state $h_{t-1}$ (carrying short-memory information), where the latter two inputs are both inherited from the previous steps.
The outputs of the network are $v_t$,  
$h_{t}$ and $c_t$.
We formally write this neural network model as,  
\begin{equation}
(v_{t}, h_{t}, c_t) =\Xi_{\vartheta}\left(g_{t-1}, h_{t-1}, c_{t-1}\right), \label{eq:rmaXi}
\end{equation}
where $\vartheta$ is the neural network parameters, and leave the details of it in  \ref{sec:lstm}. 
As one can see this network integrates the current gradient $g_{t-1}$ and the information from previous step $h_{t-1}$ and $c_{t-1}$
to produce the velocity $v_t$ that is used in DuNets.  
Finally we note that other RNN models such as the Gated Recurrent Unit (GRU)~\cite{cho2014learning} can also be used here.
In our numerical experiments we have tested both LSTM and GRU, and found no significant
difference between the performances of the two approaches, which is consistent with some existing works, e.g.,~\cite {chung2014empirical,greff2016lstm}. 
As such, we only report our experimental results with LSTM-RNN in this work.

\subsection{LPGD and LPD with MA}
Inserting the MA module into the DuNets algorithms is rather straightforward. In this section, we use LPGD and LPD as examples,
while noting that they can be implemented in other DuNets in a similar manner. 

\begin{figure}[!ht]
    \centering
    \includegraphics[width=0.85\linewidth]{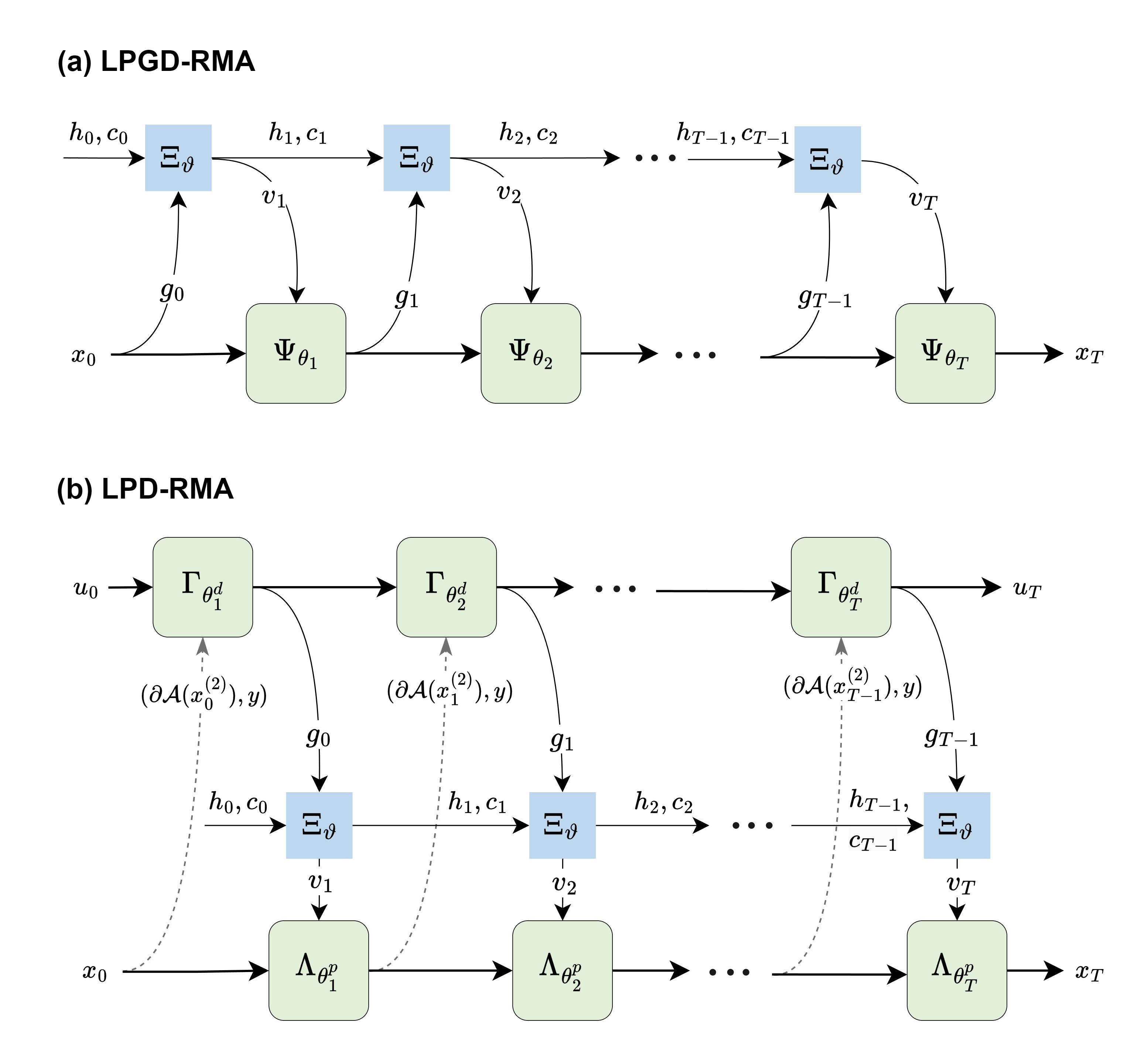}
        \caption{
        The model architectures of DuNets-RMA. The RMA module is constructed 
        via a deep LSTM-RNN, denoted as $\Xi_{\vartheta}$.
        }
    \label{fig:model} 
\end{figure}

\paragraph{LPGD}
The explicit MA can be easily incorporated with LPGD.
A notable difference is that in the unrolling methods, $g_t$ is not the gradient of the objective function, which 
may be nondifferentiable or not explicitly available. 
In LPGD, $g_t$ is taken to be the gradient of the data fidelity term only, i.e.,
 $g_t = \nabla_{x_{t-1}} \mathcal{D}(\mathcal{A}(x_{t-1}), y)$. 
The main idea here is to replace $g_{t-1}$ in Alg.~\ref{alg:lpgd} by $v_t$ calculated via Eq.~\eqref{eq:vtma}, yielding the LPGD-MA method~(Alg.~\ref{alg:lpgd-ma}).
Similarly by inserting the LSTM model into Alg.~\eqref{alg:lpgd}, we obtain the LPGD-RMA algorithm (Alg.~\ref{alg:lpgd-rma}), as illustrated in Figure~\ref{fig:model}(a).
Finally, recall that LPGD also has a shared-weights version (referred to as LPGDSW), and correspondingly we have LPGDSW-MA and LPGDSW-RMA, which will also be tested in our numerical experiments. 

\begin{algorithm}[H]
      \small
      \caption{LPGD-MA}
      \label{alg:lpgd-ma}
        \begin{algorithmic}[1]
           \item[] \textbf{Input:} $x_0 \in X, \, v_0=0$
           \item[] \textbf{Output:} $x_T$  
           \For{$t=1,\ldots,T$}
             \State{$g_{t-1} = \nabla_{x_{t-1}} \mathcal{D}(\mathcal{A}(x_{t-1}), y)$}
             \State{$v_t = \gamma v_{t-1} - \eta g_{t-1}$}
             \State{$x_t = \Psi_{\theta_t}(x_{t-1}, v_{t})$}
           \EndFor
        \end{algorithmic}
\end{algorithm}
\begin{algorithm}[H]
            \small
          \caption{LPGD-RMA algorithm}
          \label{alg:lpgd-rma}
           \begin{algorithmic}[1]
               \item[] \textbf{Input:} $x_0 \,\in X, h_0=0, c_0=0$
               \item[] \textbf{Output:} $x_T$  
               \For{$t=1,\ldots,T$}
               \State{$g_{t-1} = \nabla_{x_{t-1}} \mathcal{D}(\mathcal{A}(x_{t-1}), y)$}
                 \State{
                    $(v_{t},h_t,c_t) = \Xi_{\vartheta}(g_{t-1},h_{t-1},c_{t-1})$}
                 \State{$x_t = \Psi_{\theta_t}(x_{t-1}, {\color{black}{{v}_{t}}})$
                 }
               \EndFor
            \end{algorithmic}
    \end{algorithm}

\paragraph{LPD}
The integration of the MA schemes and LPD is a bit different.
Namely, in LPGD, the velocity is constructed based on the gradient of the data fidelity term, while in LPD, we build it upon $g_{t-1} = \left[\partial \mathcal{A}(x_{t-1}^{(1)})\right]^{*}u_{t}^{(1)}$. 
By inserting the explicit MA formula~\eqref{eq:vtma} into  Alg.~\ref{alg:lpd} we obtain the LPD-MA method (Alg.~\ref{alg:lpd-ma}).
The LPD-RMA method, depicted in Figure~\ref{fig:model}(b), can be constructed similarly: one simply replaces the explicit MA formula in Alg.~\ref{alg:lpd} with the RMA module Eq.~\eqref{eq:rmaXi}, and the complete algorithm is outlined in Alg.~\ref{alg:lpd-rma}.

\begin{algorithm}[H]
       \caption{LPD-MA algorithm}
       \label{alg:lpd-ma}
        \begin{algorithmic}[1]
           \item[] \textbf{Input:} $x_0 \in X^{N_{\text{primal}}}, u_0 \in Y^{N_{\text{dual}}}$  
           \item[] \textbf{Output:} $x_T^{(1)}$  
           \For {$t=1,\ldots,T$}
                \State $u_{t} =\Gamma_{\theta_{t}^{d}}\left(u_{t-1}, \mathcal{A}(x_{t-1}^{(2)}), y\right)$
                \State $g_{t-1} = \left[\partial \mathcal{A}(x_{t-1}^{(1)})\right]^{*}u_{t}^{(1)}$
                \State $v_t = \gamma v_{t-1} - \eta g_{t-1}$
                \State $x_{t} =\Lambda_{\theta_{t}^{p}}\left(x_{t-1}, v_t\right)$
          \EndFor
        \end{algorithmic}
\end{algorithm}

\begin{algorithm}[H]
       \caption{LPD-RMA algorithm}
       \label{alg:lpd-rma}
        \begin{algorithmic}[1]
           \item[] \textbf{Input:} $x_0 \in X^{N_{\text{primal}}}, u_0 \in Y^{N_{\text{dual}}}, h_0=0, c_0=0$ 
           \item[] \textbf{Output:} $x_T^{(1)}$  
           \For{$t=1,\ldots,T$}
             \State $u_{t} =\Gamma_{\theta_{t}^{d}}\left(u_{t-1}, \mathcal{A}(x_{t-1}^{(2)}), y\right)$
             \State $g_{t-1} = \left[\partial \mathcal{A}(x_{t-1}^{(1)})\right]^{*} (u_{t}^{(1)})$
             \State $(v_{t}, h_{t}, c_t) =\Xi_{\vartheta}\left(g_{t-1}, h_{t-1}, c_{t-1}\right)$
             \State $x_{t} =\Lambda_{\theta_{t}^{p}}\left(x_{t-1}, v_t\right)$
           \EndFor
        \end{algorithmic}
\end{algorithm}

\section{Experiments and Results}\label{sec:exp}
In this section, we present our numerical experiments on two nonlinear inverse problems: a nonlinear deconvolution and an EIT image reconstruction.

\subsection{Implementation details}
To make a fair comparison, for various DuNet methods, we adjust the number of unrolled iterations to ensure that all
the methods have approximately the same of number of training parameters. 
Specifically, for the LPGD-type of methods, we set the unrolling iterations to 20 for  LPGD-RMA  and to 43 for both LPGD and LPGD-MA. The outputs of the proximal operator unit are first concatenated with the estimated direction from the RMA module and then combined using a convolutional layer with a $3 \times 3$ kernel size and $32$ output channels before being fed to the subsequent block. 
The primal subnetwork consists of two convolutional layers of kernel size $3\times3$ and 32 output channels. 
The convolutional layers are followed by a parametric rectified linear units (PReLU) activation function.  The output convolutional layer is designed to match a desired number of channels and does not include any nonlinear activation function.
In LPD-RMA, we use 10 unrolling iterations, while in other LPD methods, we adjust it to 22.
The number of data that persists between the iterates be $N_{primal}=5, N_{dual}=5$. 
The primal subnetwork $\Gamma_{\theta_i^d}$ is the same as that used in LPGD-based methods. The dual subnetwork consists of one convolutional layer of kernel size $3\times3$ and output channels 32, and the other setting is the same as the primal subnetwork. 

All networks are trained end-to-end using Adam optimizer~\cite{adam2015} to minimize the empirical loss~\eqref{eq:mseloss}. We use a learning rate schedule according to the cosine annealing, \textit{i.e.}, the learning rate in step $t$ is
$$
\zeta_{t}=\frac{\zeta_{0}}{2}\left(1+\cos \left(\pi \frac{t}{t_{\max }}\right)\right),
$$
where the initial learning rate $\zeta_{0}$ is set to be $10^{-3}$. We also let the parameter $\beta_{2}$ of the ADAM optimizer~\cite{adam2015} to be $0.99$ and keep all other parameters as default. We perform global gradient norm clipping~\cite{zhanggradient}, limiting the gradient norms to 1 to improve training stability. We use a batch size of 32 for the nonlinear convolution example and 1 for the EIT problem. 
For the DuNets-MA methods, we choose $\gamma =0.9$  and $\eta=10^{-3}$. 
We train all models with 20 epochs and keep a set of trainable parameters that achieve minimal validation losses.
We do not enforce any constraint on the trainable parameters during training.

All experiments are run on an Intel Xeon Golden 6248 CPU and an NVIDIA Tesla V100 GPU. The nonlinear deconvolution example is run entirely on the GPU. The forward and adjoint operators in 
the EIT experiments are run on the CPU  as the \texttt{pyEIT} toolbox used is not computationally parallelizable and runs faster on the CPU. The training time for a single epoch is approximately 4 minutes in the nonlinear convolution example with 10000 training samples, and 60 minutes for the EIT example with 400 training samples. The code for implementing all experiments is available at \url{https://github.com/zhouqp631/DuNets-RMA.git}.

We use the $\ell_{2}$ loss function on the outputs from all the stages. Specifically, given the paired samples $\{x_i,y_i\}, i=1,\ldots,N$, the training objective is defined as:
\begin{equation}\label{eq:mseloss}
\mathrm{L}(\Theta)=\frac{1}{N} \sum_{i=1}^{N} \left\|\hat{x}_i-x_{i}\right\|^{2}.
\end{equation}
Here, $\hat{x}_i$ is the reconstruction, and $\Theta$ presents the set of trainable parameters.

\subsection{A nonlinear deconvolution problem}\label{sec:exp-deconv}
\subsubsection{Problem setting}
 We consider a nonlinear deconvolution problem 
which is constructed largely following~\cite{zoumpourlis2017non}. 
For each input $\mathbf{x}=[x_{1}, x_{2}, \cdots, x_{n}]'$ consisting of $n$ elements, the forward problem is defined as
\begin{equation}\label{eq:ndp_forward}
y(\mathbf{x}) = a\cdot\mathbf{x}' \mathbf{W}_{2} \mathbf{x}+\mathbf{w}_{1}' \mathbf{x}+b.
\end{equation}
Here $\-w_1=[w_{1}^{1}, w_{1}^{2}, \cdots, w_{1}^{n}]'$ is the first-order Volterra kernel, which contains the coefficients of the Volterra series' linear part. The second-order Volterra kernel, denoted as {$\-W_2$}, is structured as follows:
\begin{equation*}
\mathbf{W}_{2}=\left[\begin{array}{cccc}w_{2}^{1,1} & w_{2}^{1,2} & \cdots & w_{2}^{1, n} \\ 0 & w_{2}^{2,2} & \cdots & w_{2}^{2, n} \\ \vdots & \vdots & \ddots & \vdots \\ 0 & 0 & \cdots & w_{2}^{n, n}\end{array}\right]
\end{equation*}
It is important to note that the parameter $a$ controls the degree of nonlinearity in the deconvolution problem.

\subsubsection{Training and testing datasets}
We assume that the unknown $x$ is on 53 mesh grid points, and meanwhile we choose the nonlinear kernel with size 9 and stride 4, and the dimension of the observed data $y$ is 12. The first-order and second-order Volterra kernels in Eq.~\eqref{eq:ndp_forward} are derived using the methods described in Section 3 of~\cite{kumar2009volterrafaces}.
We consider four sets of experiments with different coefficients in Eq.~\eqref{eq:ndp_forward}: $a=0, 1, 2, 4$. We generate the ground truth by sampling from a TV prior (see Chapter 3.3 in~\cite{kaipio2006statistical}) and then obtain the observation data by~\eqref{eq:ndp_forward}. We employ 12000 randomly generated paired samples, where 10000 pairs are used as the training set, and the remaining 1000/1000 pairs are used as the validation/test sets.

\subsubsection{Results and discussion}
\begin{figure}[!htb]
\centering
  \begin{subfigure}[t]{0.43\textwidth}
    \centering
    \includegraphics[width=0.9\linewidth]{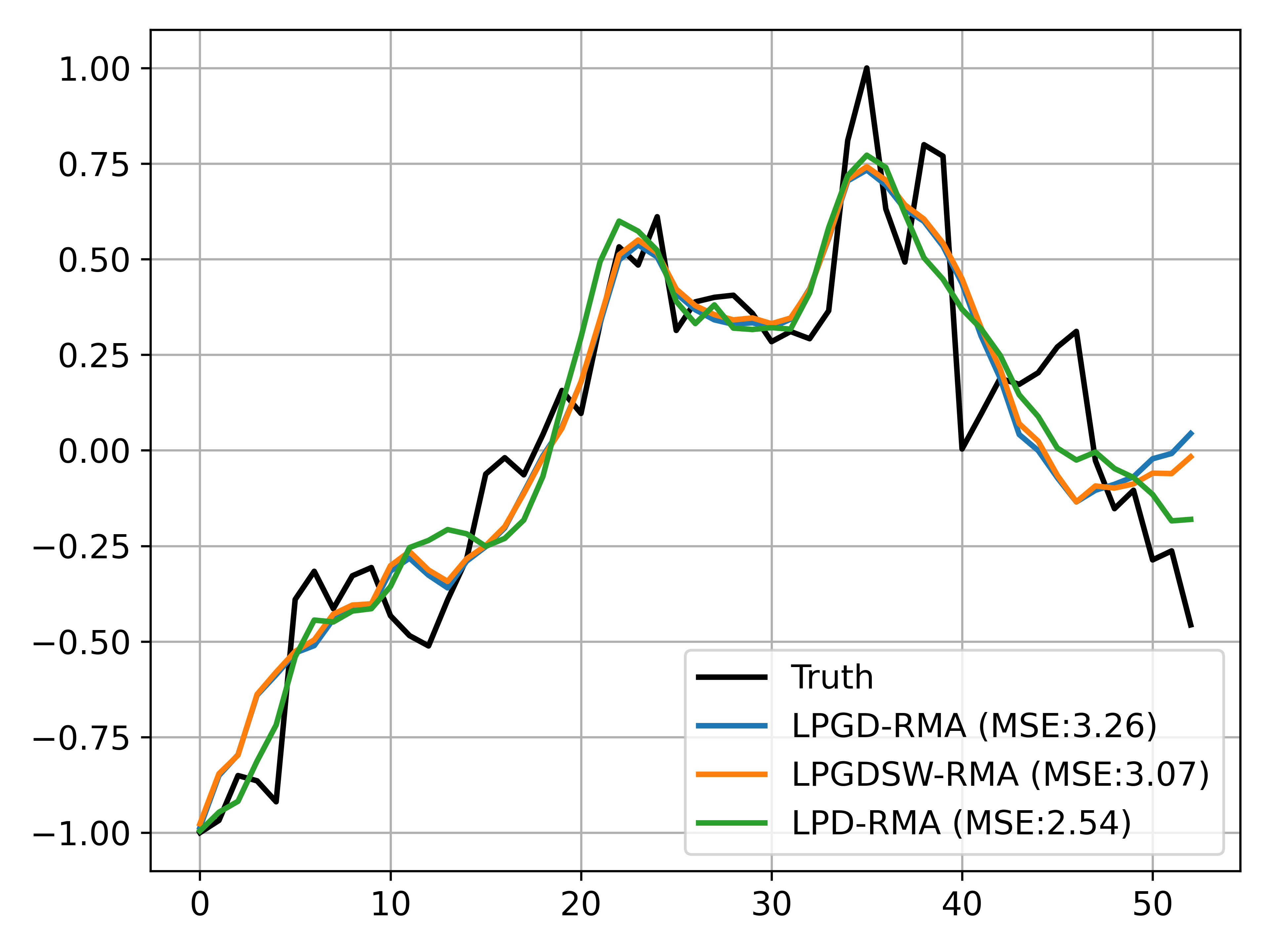}
    \caption{$a=0$}
  \end{subfigure}
  \begin{subfigure}[t]{0.43\textwidth}
    \centering
    \includegraphics[width=0.9\linewidth]{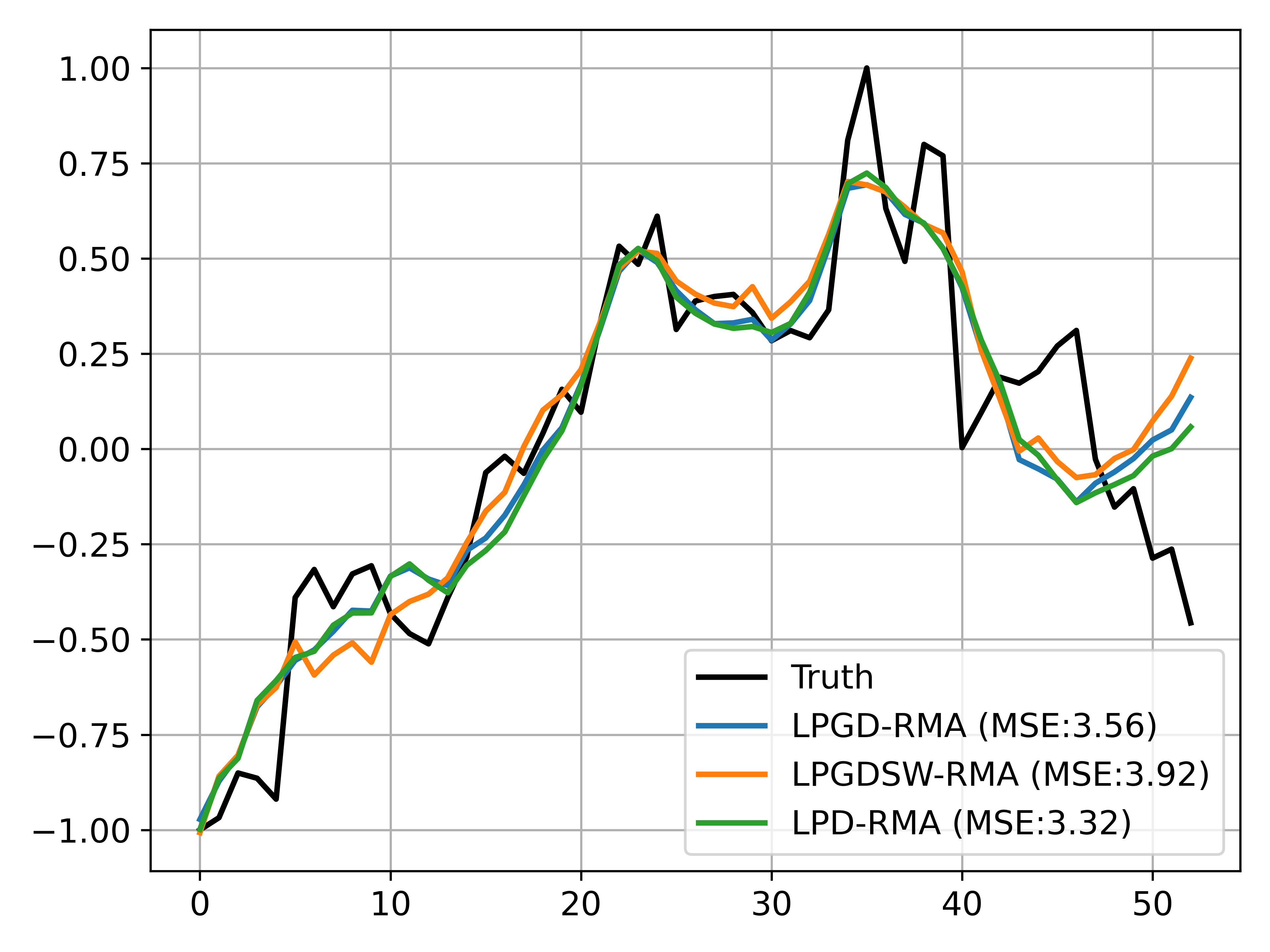}
    \caption{$a=1$}
  \end{subfigure}

  \begin{subfigure}[t]{0.43\textwidth}
    \centering
    \includegraphics[width=0.9\linewidth]{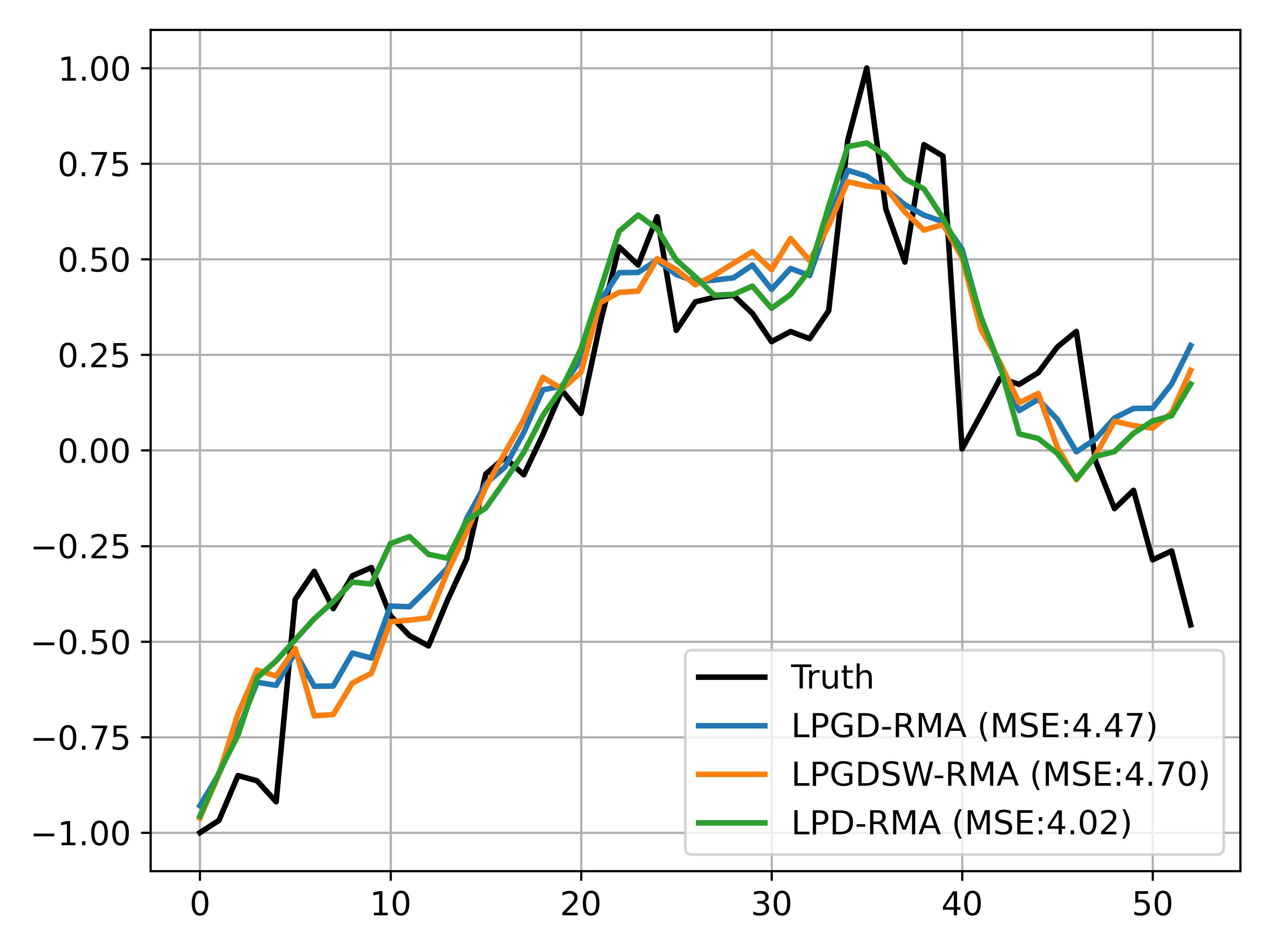}
    \caption{$a=2$}
  \end{subfigure}
    \begin{subfigure}[t]{0.43\textwidth}
    \centering
    \includegraphics[width=0.9\linewidth]{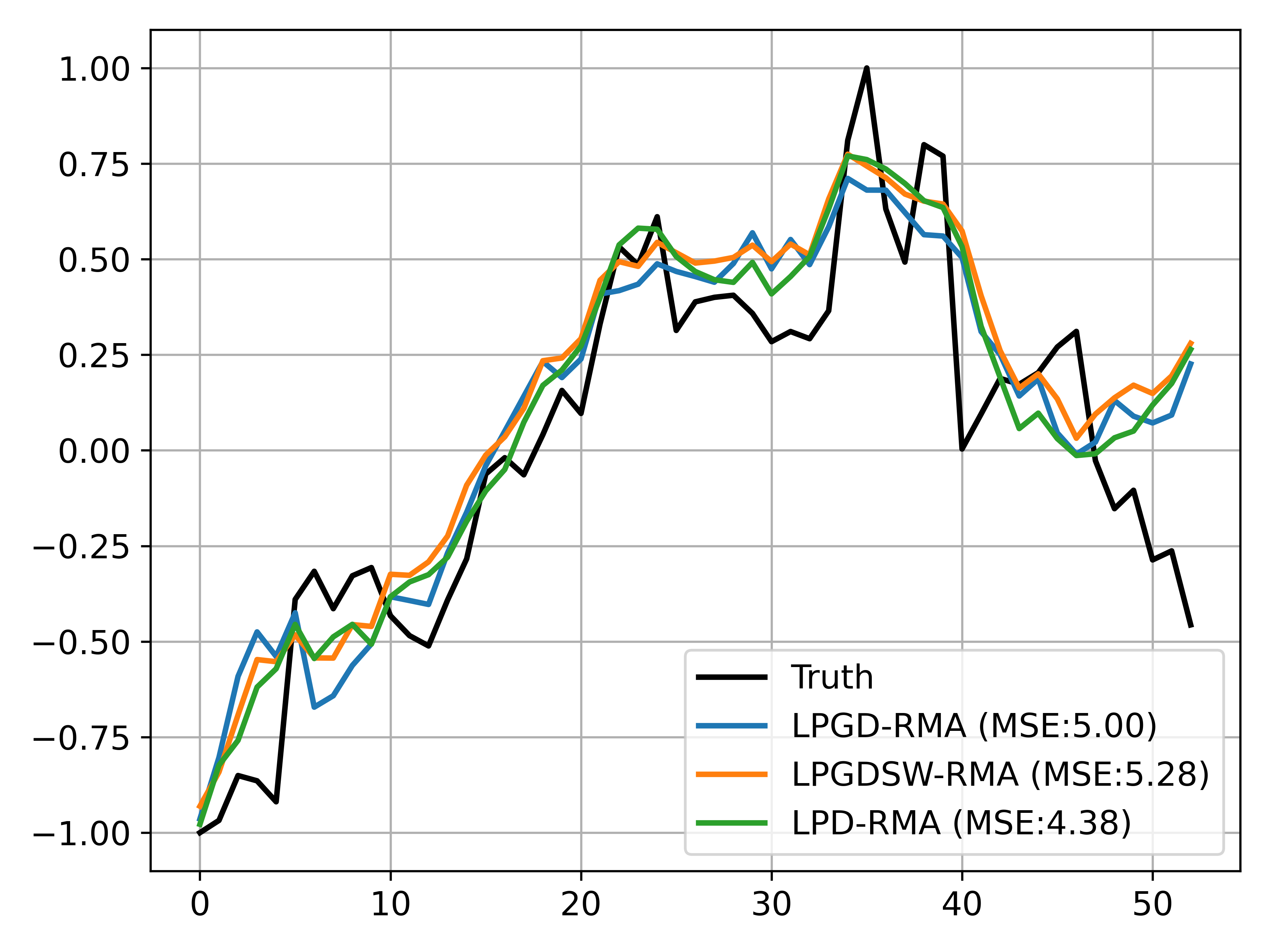}
    \caption{$a=4$}
  \end{subfigure}
 \caption{Deconvolution results and their corresponding MSE values for all DuNets-RMA methods.}
  \label{fig:toy_reconstructs}
\end{figure}

\paragraph{Benefit of the RMA scheme}
We assess the performance of RMA with the three unrolling methods LPGDSW, LPGD and LPD, and in each method we implement the following three different cases: without acceleration, with the conventional MA module and with the RMA module;
as such there are 9 schemes implemented in total. 
All hyper-parameters in the tested methods are manually tuned for optimal performance or automatically chosen as described in the aforementioned references. Table~\ref{tab:toy-results} demonstrates the performance of each method in terms of the mean-square error (MSE) on four different settings: $a=0,1,2,4$.  The visual comparison can be found in Figure~\ref{fig:toy_reconstructs}. 
We summarize our findings as the following: 
\begin{enumerate}[label=(\roman*)]
  \item when $a=0$, the MSE values of each type of DuNets method are almost the same,
  which is not surprising as  the gradient of the forward operator is constant;
  \item when $a > 0$, DuNets with RMA outperform the state-of-the-art methods by a rather 
  large margin (\textit{e.g.}, LPD-RMA outperforms the LPD method by 8.0\%, 12.0\%, and 16.0\% in terms of MSE for $a=1, 2, 4$ respectively), suggesting that  the RMA module can
  significantly improve the performance of DuNets, especially for problems 
  that are highly nonlinear;
  \item the conventional MA module can also improve the performance, but RMA clearly outperforms it in all the nonlinear cases; 
  \item LPD-RMA method consistently achieves the best results in terms of MSE.
\end{enumerate}

\begin{table}[!htb]
\centering
\caption{MSE results of the DuNets methods under different $a$ values. 
The result of  LPGD is not reported as it fails to converge. The best results are indicated in \textcolor{orange}{orange} color.}
\label{tab:toy-results}
\begin{tabular}{lclclcll}
\hline
          & $a=0$        &  & $a=1$        &  & $a=2$        &  & \multicolumn{1}{c}{$a=4$} \\ \hline \specialrule{0em}{0pt}{3pt}
LPGD       &   ---      &  &   ---      &  &     ---    & &   ---                 \\ \specialrule{0em}{3pt}{3pt}
LPGD-MA    &  3.21E-02        &  &  5.37E-02        &  &  6.49E-02        &  &  7.76E-02                     \\ \specialrule{0em}{3pt}{3pt}
LPGD-RMA   &  3.23E-02        &  &  \textcolor{orange}{3.56E-02}        &  &  \textcolor{orange}{4.43E-02}        &  &  \textcolor{orange}{4.97E-02}                     \\ \hline  
\specialrule{0em}{0pt}{3pt}

LPGDSW     &  3.01E-02        &  &   4.61E-02       &  &  5.88E-02        &  & 6.85E-02                      \\ \specialrule{0em}{3pt}{3pt}
LPGDSW-MA  &  3.02E-02        &  &   4.54E-02       &  &  5.32E-02        &  & 5.78E-02                      \\ \specialrule{0em}{3pt}{3pt}
LPGDSW-RMA &  3.01E-02        &  &   \textcolor{orange}{3.89E-02}        &  &  \textcolor{orange}{4.68E-02} &  &   \textcolor{orange}{5.24E-02}                       \\ \hline  \specialrule{0em}{0pt}{3pt}

LPD       & 2.69E-02   &  & 3.65E-02 &  & 4.61E-02 &  & 5.17E-02              \\ \specialrule{0em}{3pt}{3pt}
LPD-MA    & 2.68E-02 &  &  3.71E-02 & &4.65E-02 & &5.22E-02                              \\ \specialrule{0em}{3pt}{3pt}
LPD-RMA   & 2.67E-02 &  & \textcolor{orange}{3.35E-02} &  & \textcolor{orange}{4.04E-02}  &  & \textcolor{orange}{4.33E-02}             \\ \hline  \specialrule{0em}{0pt}{3pt}
\end{tabular}
\end{table}

Next we will test the sensitivity of the RMA module with respect to both the network structure and the data size. 
Since the LPGD-type methods are significantly outperformed by the LPD-type ones in this example, we only use the LPD-type methods in these tests.

\paragraph{Sensitivity to the RMA structure}
We discuss here the choice for the architecture of the RMA modules, \textit{i.e.}, the 
number of hidden layers $L$ and the hidden size $n$ of the LSTM layer.  
To avoid overfitting, we limit the ranges of the hidden layers as $L \in \{1,2,3\}$ and the hidden size as $n \in \{30, 50, 70\}$. 
Table~\ref{tab:different-L-toy} demonstrates the results of LPD-RMA trained with different network structures in the setting where $a=1$. 
\begin{table}[!htb]
\centering
\caption{Mean MSE values of the LPD-RMA models with $L=1,2,3$ and $n=30, 50, 70$. Evaluation is done via repeating the experiment 10 times. The number of trainable parameters is also reported below the MSE value in parentheses.}
\label{tab:different-L-toy}
\begin{tabular}{c|lll}
\hline
\multicolumn{1}{l|}{\diagbox{$n$}{$L$}} & \multicolumn{1}{c}{1} & \multicolumn{1}{c}{2}        & \multicolumn{1}{c}{3}        \\ \hline
30      
& \begin{tabular}[c]{@{}c@{}}3.37E-02\\(94193)\end{tabular}  
& \begin{tabular}[c]{@{}c@{}}3.36E-02\\(101363)\end{tabular} 
& \begin{tabular}[c]{@{}c@{}}3.34E-02\\(108533)\end{tabular}   \\ \specialrule{0em}{3pt}{3pt}
50      
& \begin{tabular}[c]{@{}c@{}}3.35E-02\\(103353)\end{tabular} 
& \begin{tabular}[c]{@{}c@{}}3.34E-02\\(121303)\end{tabular}   
& \begin{tabular}[c]{@{}c@{}}3.33E-02\\(139253)\end{tabular}   \\ \specialrule{0em}{3pt}{3pt}
70     
& \begin{tabular}[c]{@{}c@{}}3.34E-02\\(114913)\end{tabular}   
& \begin{tabular}[c]{@{}c@{}}3.35E-02\\(148443)\end{tabular} 
& \begin{tabular}[c]{@{}c@{}}3.32E-02\\(181973)\end{tabular}   \\ 
\hline
\end{tabular}
\end{table}
We observe that in all these settings the LPD-RMA yields similar results, 
indicating that the algorithm is rather robust provided that the parameter values are within a reasonably defined range. 
With extensive numerical tests, we have found that a reasonable choice of $L$ may be $L\in\{1, 2\}$ in moderate dimensions, and 
$n$ can be chosen to be approximately the same as the dimensionality of the unknown variables. We have also tested the methods for $a=2, 4$, where the results are qualitatively similar to those shown in Table~\ref{tab:different-L-toy}, 
and hence are omitted here.

\begin{figure}[!htb]
\centering
\includegraphics[width=0.5\linewidth]{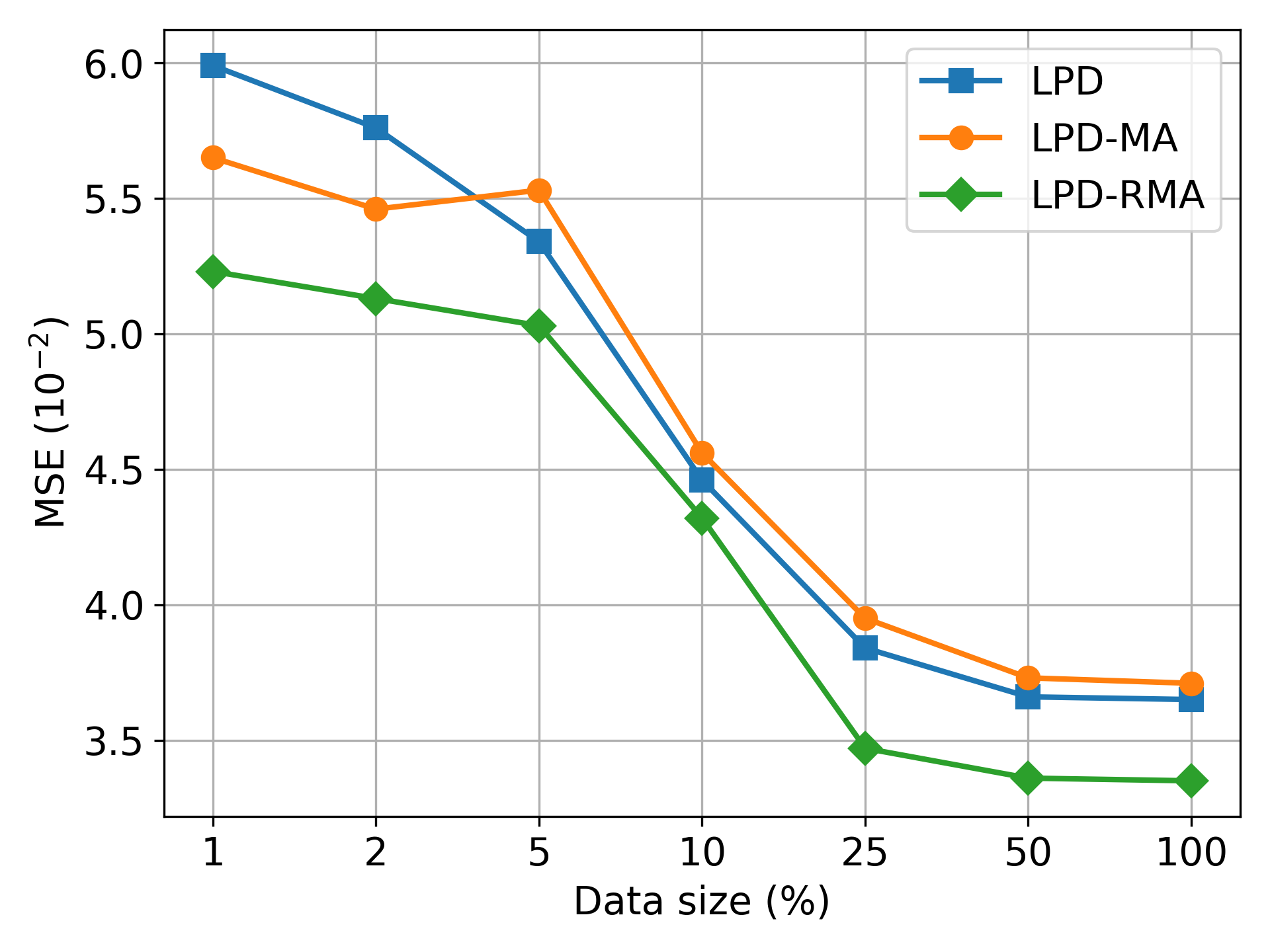}
\caption{The MSE results plotted against the data size for $a=1$.}
\label{fig:toy-differnt-datasize}
\end{figure}

\paragraph{Sensitivity to data size}
To assess the data efficiency of the proposed methods, we train them on different data sizes.
The data size is measured as the percentage of the total available training data, and 
the MSE results are plotted against it in Figure~\ref{fig:toy-differnt-datasize}. 
The figure shows that LPD-RMA is considerably more data-efficient than LPD and LPD-MA. 
Interestingly after the data size increases to over 50\%, the use of the conventional MA module
can not improve the performance of LPD, while LPD-RMA consistently achieves the best accuracy across the whole range.

\subsection{Electrical impedance tomography}\label{sec:exp-eit}
\subsubsection{Problem setting}
EIT is a nondestructive imaging technique that aims at reconstructing the inner conductivity distribution of a medium from a set of voltages registered on the boundary of the domain by a series of electrodes~\cite{zou2003review}. 

In this example, we consider a bounded domain $\Omega \subseteq \mathbb{R}^{2}$ with a boundary $\partial \Omega$ containing certain conducting materials whose electrical conductivity is defined by a positive spatial function $\sigma(x) \in L^{\infty}(\Omega)$.
Next, we assume that $L$ different electrical currents are injected into the boundary of $\partial \Omega$, and the resulting electrical potential should satisfy the following  governing equations with the same coefficient but different boundary conditions:
\begin{equation}\label{eq:eit_pde}
\left\{\begin{array}{ll}\nabla \cdot(\sigma \nabla u)=0 & \text { in } \Omega \\ 
u+z_{l} \sigma \frac{\partial u}{\partial e}=V_{l} & \text { on } E_{l},~~ l=1, \ldots, L \\ 
\int_{E_{l}} \sigma \frac{\partial u}{\partial e} \mathrm{~d} s=I_{l} & \text { on } \Gamma \\ 
\sigma \frac{\partial u}{\partial e}=0 & \text { on } \tilde{\Gamma}
\end{array}\right.
\end{equation}
where $\Gamma(\tilde{\Gamma})$ is the boundary $\partial \Omega$ with (without) electrodes, $e$ is the outer normal direction at the boundary, $V_{l}$ is the voltage measured by the $l$-th electrode $E_{l}$ when current $I_{l}$ is applied, and $z_{l}$ is the contact impedance.

Numerically, the object domain $\Omega$ discretized into $n_{S}$ subdomains $\left\{\tau_{j}\right\}_{j=1}^{n_{S}}$ and
$\sigma$ is constant over each of them. 
One injects a current at a fixed frequency through a pair of electrodes attached to the boundary and measures the voltage differences on the remaining electrode pairs. This process is repeated over all electrodes, and the resulting data is represented as a vector denoted by $y \in \mathbb{R}^{n_Y}$ where $n_Y$ is the number of measurements.
We can define a mapping $F: \mathbb{R}^{n_S} \rightarrow \mathbb{R}^{n_Y}$
representing the discrete version of the forward operator:
\begin{equation}\label{eq:eit_nonlinear}
y = F(\sigma)+\eta, 
\end{equation}
where $\eta$ is a zero-mean Gaussian-distributed measurement noise.
The EIT problem aims to estimate the static conductivity $\sigma$ from measurements $y$.
The EIT problem is widely considered to be challenging due to its severe ill-posedness, largely caused by the highly non-linear dependence of the boundary currents on the conductivity.

\subsubsection{Training and Testing Datasets}\label{sec:eit-dataset}
We run numerical tests on a set of synthetic $2 \mathrm{D}$ experiments to evaluate the performance of the reconstruction methods. 
In the circular boundary ring, $L=16$ electrodes are equally spaced and located. The conductivity of the background liquid is set to be $\sigma_{0}=1.0 \Omega m^{-1}$. Measurements are simulated by using \texttt{pyEIT}~\cite{liu2018pyeit},
a Python-based package for EIT. 
For each simulated conductivity phantom, the forward EIT problem~\eqref{eq:eit_pde} is solved using FEM with approximately 1342 triangular elements. 
We explore the following two typical cases to test the methods:
\begin{itemize}
\item Case 1: the anomalies consist of two random circles with radii
generated from the uniform distribution ${U}(-0.6,0.6)$ and the conductivity values are 0.5 and 2 respectively in each circle; 

\item Case 2: the anomalies consist of four random circles with  radii generated according to ${U}(-0.55, 0.55)$ 
and the conductivity values are 0.3, 0.5, 1.5, and 2.0.
\end{itemize}
We perform the DuNet methods with three different training sample sizes 50, 200 and 400, and 20 testing samples to evaluate the performance of the methods.
We report that the testing time is approximately 19 seconds per sample for all DuNets.

\subsubsection{Results and discussion}
In the numerical experiments, we use the same set of unrolling schemes as in Section~\ref{sec:exp-deconv}, and in addition we also implement the regularized Gauss-Newton (GN) method and the primal-dual interior point method with total variation regularizer (PDIPM-TV)~\cite{zhou2015comparison}. The parameters in both GN and PDIPM-TV are optimally tuned. 
We calculate the mean and standard deviation of MSE over ten independent runs with different training data sizes, 
and provide the results for Case 1 in Table~\ref{tab:eit2targets} and those for Case 2 in Table~\ref{tab:eit-4-targets}.
In what follows our discussion is focused on three aspects of the experimental results, 
\begin{enumerate*}[label=(\roman*)]
  \item benefit of the RMA module,
  \item behaviors in low-data regimes, 
  \item robustness against the number of inclusions in the conductivity area,
and 
  \item performance against the number of unrolling iterations.
\end{enumerate*}

\begin{figure}[!ht]
    \centering
    \includegraphics[width=0.95\linewidth]{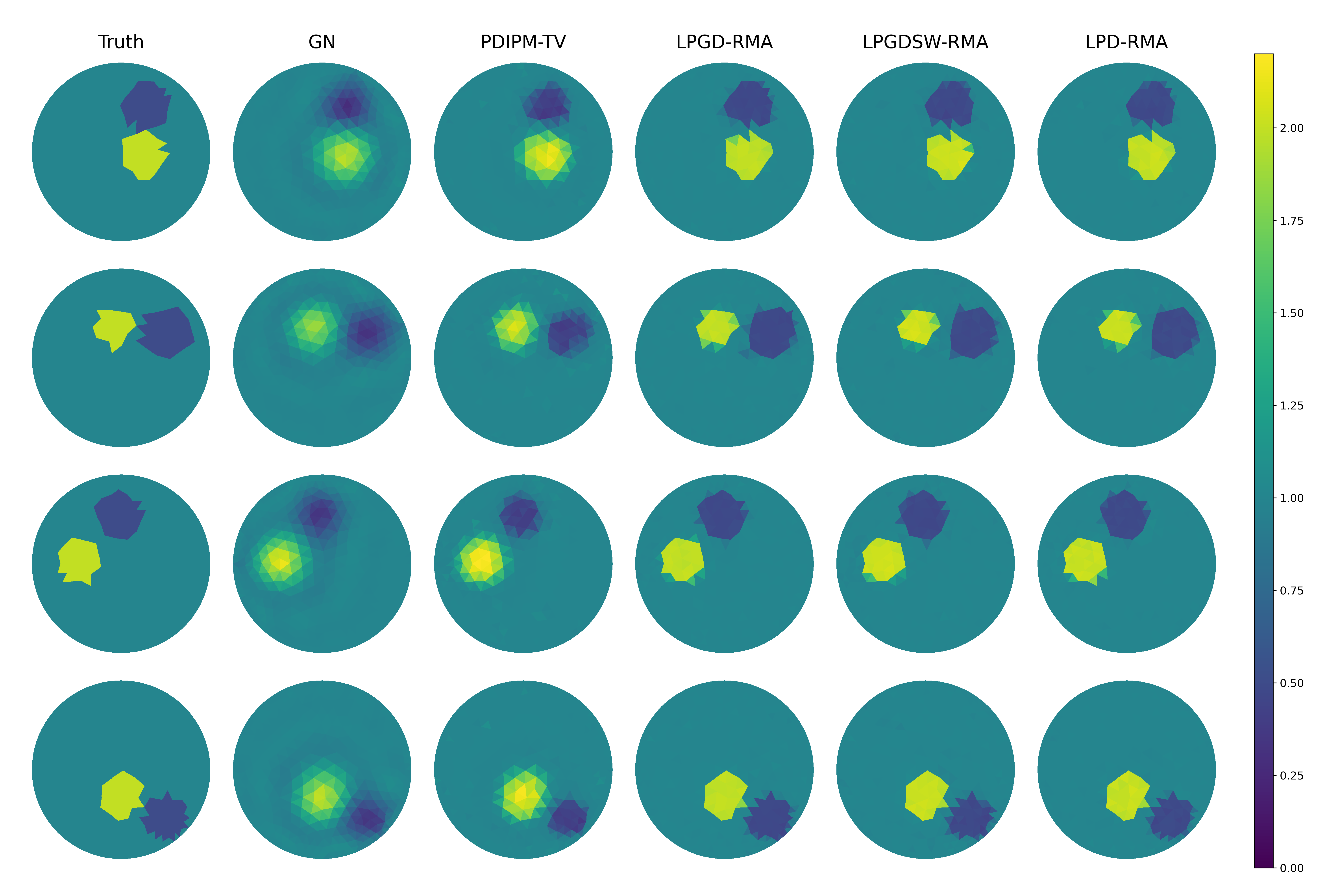}
    \caption{EIT reconstruction results of four testing samples
    in Case 1 (2 inclusions). From left to right: ground truth, GN, PDIPM-TV, LPGD-RMA, LPGDSW-RMA, and LPD-RMA.}
    \label{fig:eit-2-reconst}    
\end{figure}

\begin{figure}[!ht]
    \raggedleft 
    \includegraphics[width=0.98\linewidth]{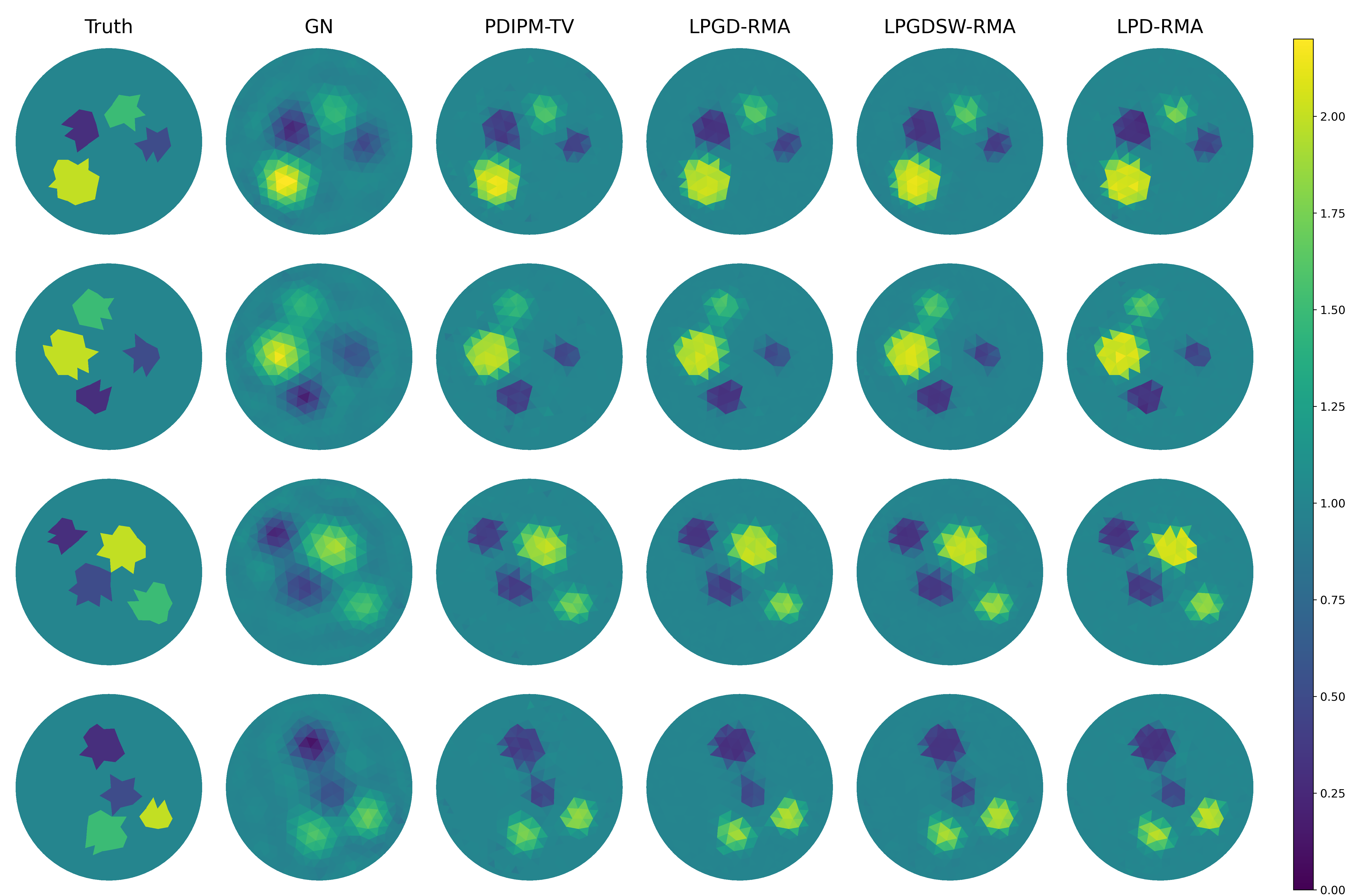}
    \caption{EIT reconstruction results of four testing samples
    in Case 2 (4 inclusions). From left to right: ground truth, GN, PDIPM-TV, LPGD-RMA, LPGDSW-RMA, and LPD-RMA.}
    \label{fig:eit-4-reconst}    
\end{figure}

\paragraph{Benefit of RMA scheme} 
First we note that five of the ten runs of the baseline LPGD (with no momentum acceleration) fail to converge within 20 epochs in both cases, and as such, we omit the results of the method in all the tables and figures. 
All the other algorithms can reasonably capture the inclusions' shape and position in all the ten runs. 
We highlight that,  according to Tables~\ref{tab:eit2targets} and~\ref{tab:eit-4-targets} the methods with the RMA module achieve the best performance in all but one test (LPGDSW method with 50 samples for Case 1) where the standard MA has the best results. 
In contrast, the effect of the standard MA module is not consistent: for example it results in worse performance than the baseline approach (without momentum acceleration) in the LPD method for case 1. 
The learning-based RMA module provides a more effective implicit regularizer than standard DuNets by utilizing previous gradient information.
As such, we can see that DuNets-RMA achieves improved and more stable performance relative to standard DuNets. 
Moreover, we want to compare the proposed methods with the conventional optimization based approach. 
To do so we provide the reconstruction results of four testing samples in Figure~\ref{fig:eit-2-reconst} (for Case 1)
and Figure~~\ref{fig:eit-4-reconst} (for Case 2). For simplicity we only provide the results of the DuNets with RMA, which is compared with those obtained by GN and PDIPM-TV methods. It can be seen from both figures that all DuNets approaches with the RMA module can yield rather accurate reconstruction for all the inclusions, and the quality of the images is clearly better than those of GN and PDIPM-TV (note that the parameters in GN and PDIPM-TV are manually tuned for the best MSE results). Furthermore, the inclusions near the boundary are better recovered than those near the center, which confirms that inclusions far away from the boundary are more difficult to reconstruct since the boundary data are not sensitive to them~\cite{yang2020gauss,guo2021construct}.

\begin{table}[!htb]
\centering
\caption{The average MSE values of the DuNets methods for Case 1, with the associated standard deviations in parentheses. 
The average MSE for the GN approach is 10.8E-03 ($\pm$ 3.16E-03), and for the PDIPM-TV method, it is 5.24E-03 ($\pm$ 2.78E-03).
The best MSE results are indicated in \textcolor{orange}{orange} color.
}\label{tab:eit2targets}
\begin{tabular}{llll}
\hline
data size   &
  \multicolumn{1}{c}{50} &
  \multicolumn{1}{c}{200} &
  \multicolumn{1}{c}{400} \\ \hline
LPGD &
  \multicolumn{1}{c}{---} &
  \multicolumn{1}{c}{---} &
  \multicolumn{1}{c}{---} \\  \specialrule{0em}{2pt}{2pt}
LPGD-MA &
  \begin{tabular}[c]{@{}c@{}}6.13E-03\\ ($\pm$ 16.1E-04)\end{tabular} &
  \begin{tabular}[c]{@{}c@{}}5.17E-03\\ ($\pm$ 14.1E-04)\end{tabular} &
  \begin{tabular}[c]{@{}c@{}}4.18E-03\\ ($\pm$ 10.5E-04)\end{tabular} \\  \specialrule{0em}{2pt}{2pt}
LPGD-RMA &
  \begin{tabular}[c]{@{}c@{}}\textcolor{orange}{3.02E-03}\\ ($\pm$ 1.34E-04)\end{tabular} &
  \begin{tabular}[c]{@{}c@{}}\textcolor{orange}{2.48E-03}\\ ($\pm$ 1.08E-04)\end{tabular} &
  \begin{tabular}[c]{@{}c@{}}\textcolor{orange}{2.25E-03}\\ ($\pm$ 1.41E-04)\end{tabular}  \\  \hline 

LPGDSW &
  \begin{tabular}[c]{@{}c@{}}4.33E-03\\ ($\pm$ 13.7E-04)\end{tabular} &
  \begin{tabular}[c]{@{}c@{}}2.87E-03\\ ($\pm$ 1.15E-04)\end{tabular} &
  \begin{tabular}[c]{@{}c@{}}3.07E-03\\ ($\pm$ 1.86E-04)\end{tabular} \\  \specialrule{0em}{2pt}{2pt}
LPGDSW-MA &
  \begin{tabular}[c]{@{}c@{}}\textcolor{orange}{3.81E-03}\\ ($\pm$ 3.15E-04)\end{tabular} &
  \begin{tabular}[c]{@{}c@{}}2.92E-03\\ ($\pm$ 1.40E-04)\end{tabular} &
  \begin{tabular}[c]{@{}c@{}}2.95E-03\\ ($\pm$ 1.55E-04)\end{tabular}  \\ \specialrule{0em}{2pt}{2pt}
LPGDSW-RMA &
  \begin{tabular}[c]{@{}c@{}}3.92E-03\\ ($\pm$ 4.79E-04)\end{tabular} &
  \begin{tabular}[c]{@{}c@{}}\textcolor{orange}{2.65E-03}\\ ($\pm$ 1.99E-04)\end{tabular} &
  \begin{tabular}[c]{@{}c@{}}\textcolor{orange}{2.63E-03}\\ ($\pm$ 1.14E-04)\end{tabular} \\ \hline
LPD &
  \begin{tabular}[c]{@{}c@{}}3.25E-03\\ ($\pm$ 1.87E-04)\end{tabular} &
  \begin{tabular}[c]{@{}c@{}}2.55E-03\\ ($\pm$ 1.34E-04)\end{tabular} &
  \begin{tabular}[c]{@{}c@{}}2.35E-03\\ ($\pm$ 2.13E-04)\end{tabular}  \\ \specialrule{0em}{2pt}{2pt}
LPD-MA &
  \begin{tabular}[c]{@{}c@{}}3.29E-03\\ ($\pm$ 1.09E-04)\end{tabular} &
  \begin{tabular}[c]{@{}c@{}}2.71E-03\\ ($\pm$ 1.46E-04)\end{tabular} &
  \begin{tabular}[c]{@{}c@{}}2.44E-03\\ ($\pm$ 1.64E-04)\end{tabular}  \\ \specialrule{0em}{2pt}{2pt}
LPD-RMA &
  \begin{tabular}[c]{@{}c@{}}\textcolor{orange}{3.11E-03}\\ ($\pm$ 1.52E-04)\end{tabular} &
  \begin{tabular}[c]{@{}c@{}}\textcolor{orange}{2.17E-03}\\ ($\pm$ 1.36E-04)\end{tabular} &
  \begin{tabular}[c]{@{}c@{}}\textcolor{orange}{2.04E-03}\\ ($\pm$ 1.89E-04)\end{tabular}  \\ \hline
\end{tabular}
\end{table}

\begin{figure}[!htb]
\centering
  \begin{subfigure}[t]{0.32\textwidth}
    \centering
    \includegraphics[width=0.95\linewidth]{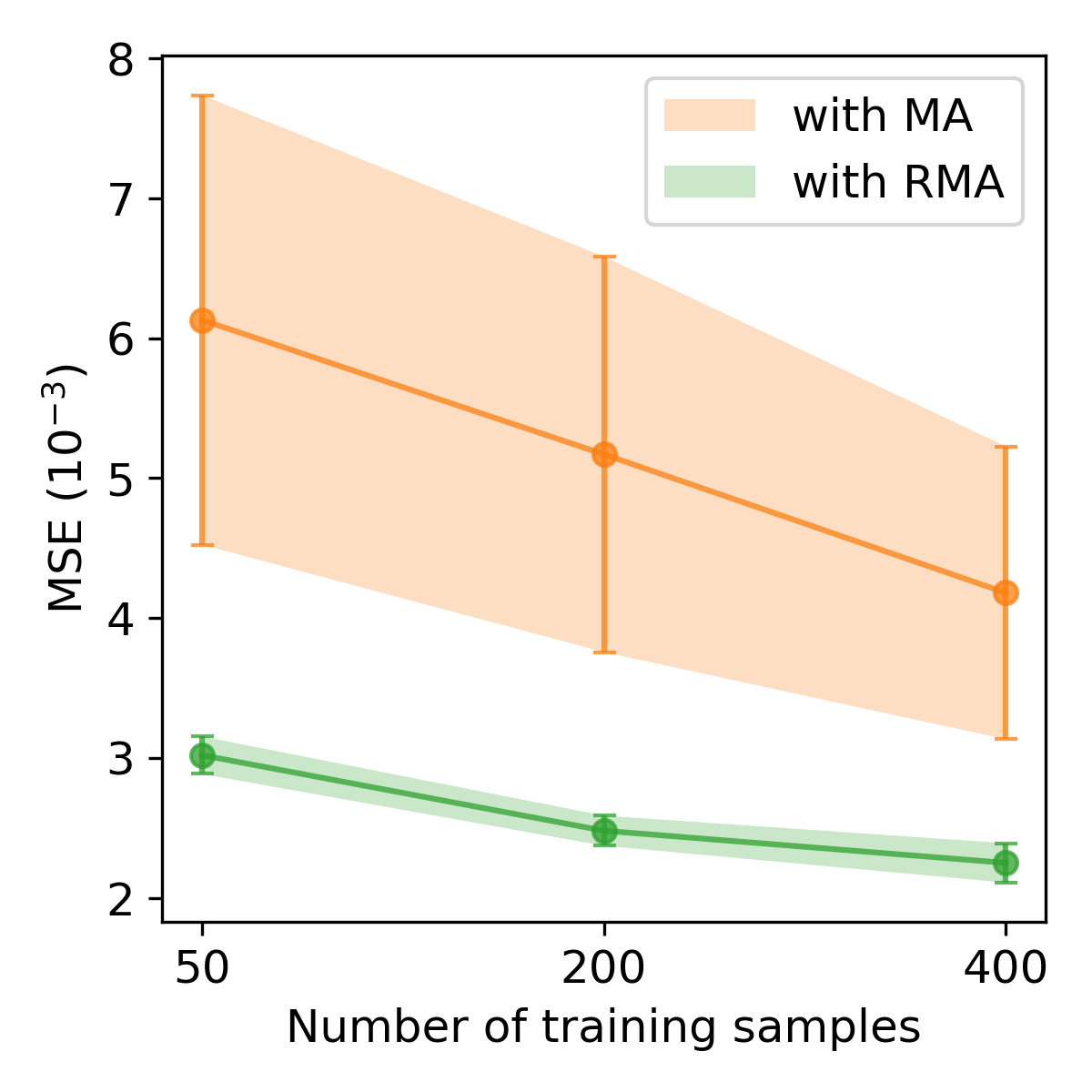}
    \caption{LPGD}
  \end{subfigure}
  \begin{subfigure}[t]{0.32\textwidth}
    \centering
    \includegraphics[width=0.95\linewidth]{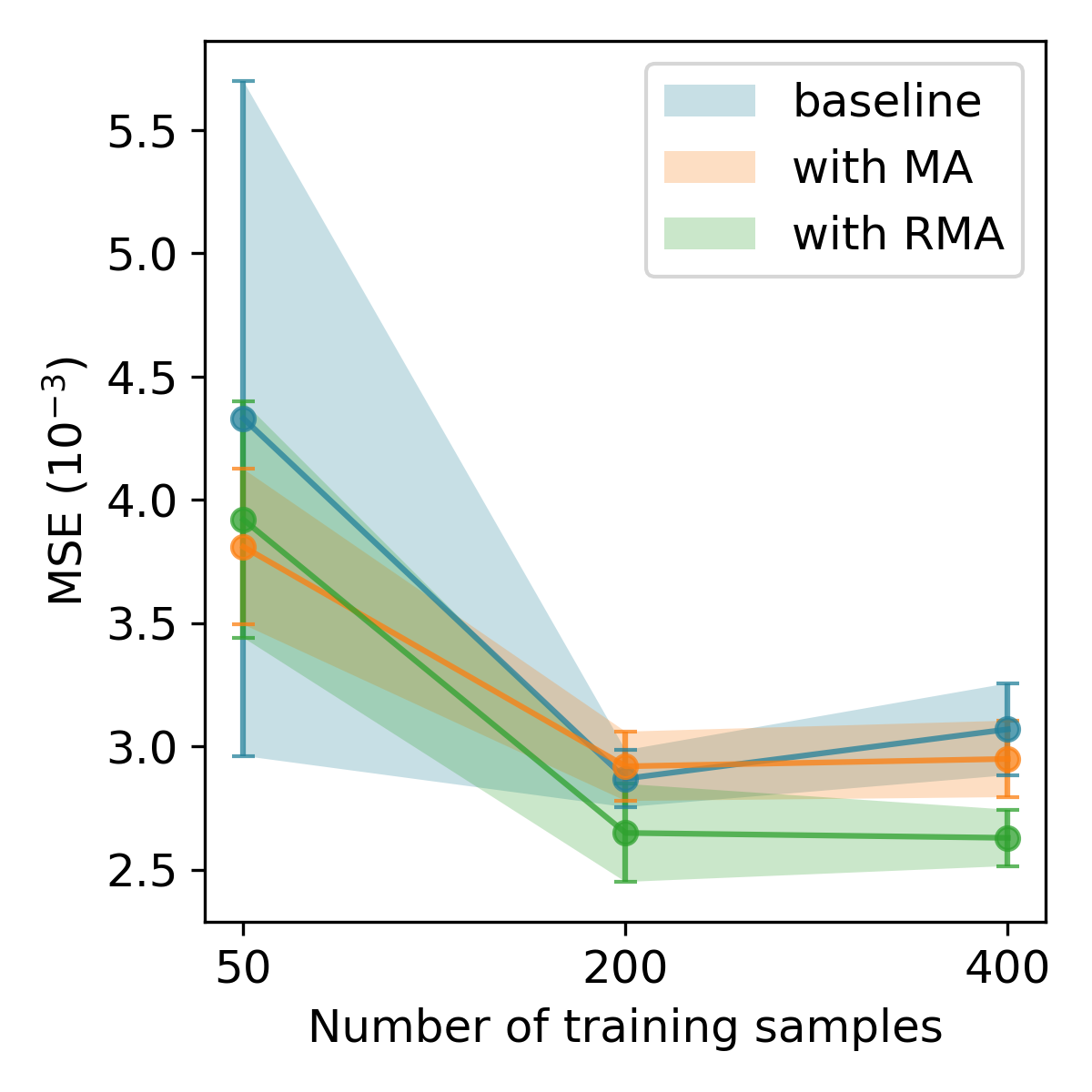}
    \caption{LPGDSW}
  \end{subfigure}
  \begin{subfigure}[t]{0.32\textwidth}
    \centering
    \includegraphics[width=0.95\linewidth]{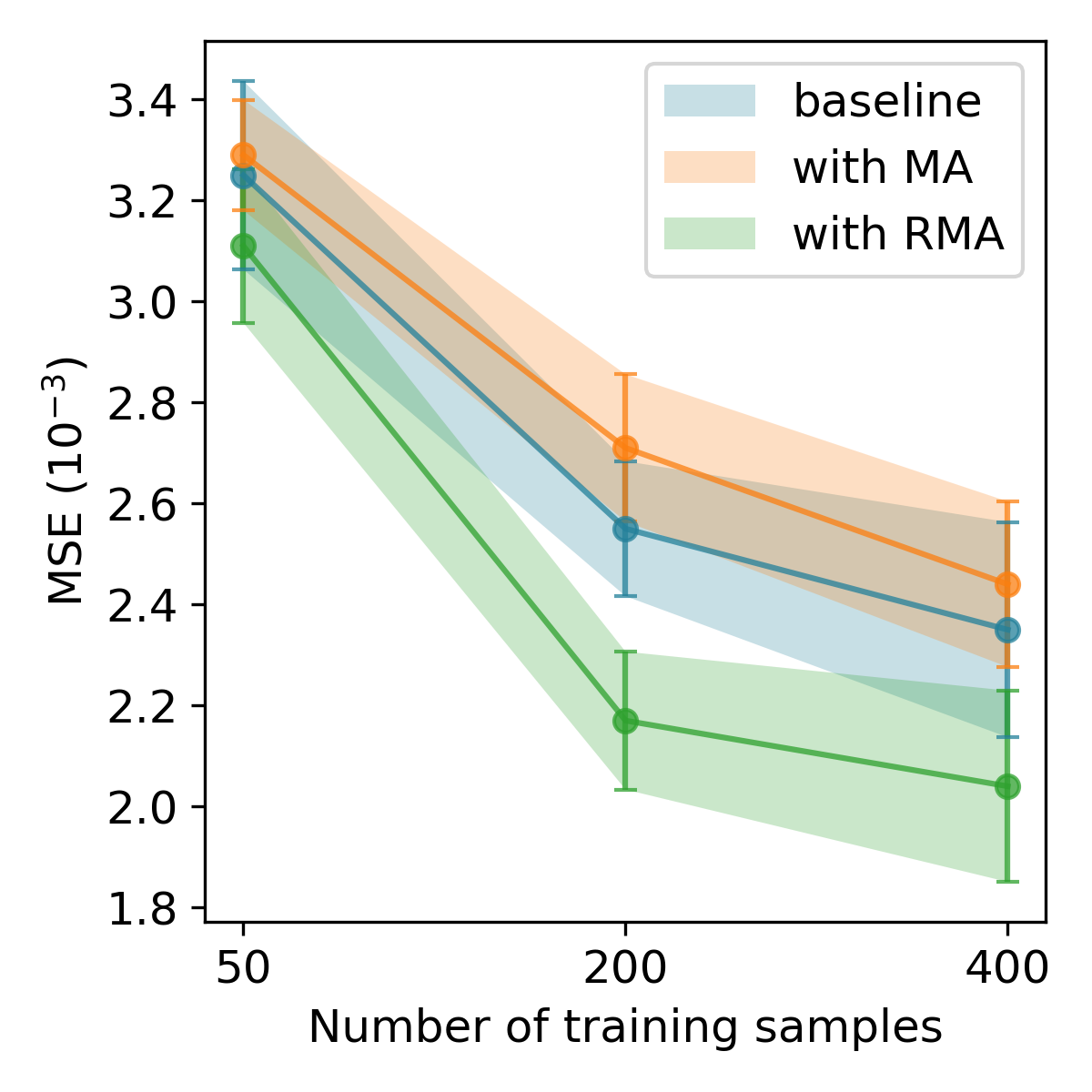}
    \caption{LPD}
\end{subfigure}
 \caption{The MSE results plotted against 
 the number of training data for Case 1. The solid line represents the average of 10 tests, and the shade around the solid line depicts the one standard deviation. }
  \label{fig:eit-2-differnt-training-number}
\end{figure}

\begin{figure}[!htb]
\centering
  \begin{subfigure}[t]{0.32\textwidth}
    \centering
    \includegraphics[width=0.95\linewidth]{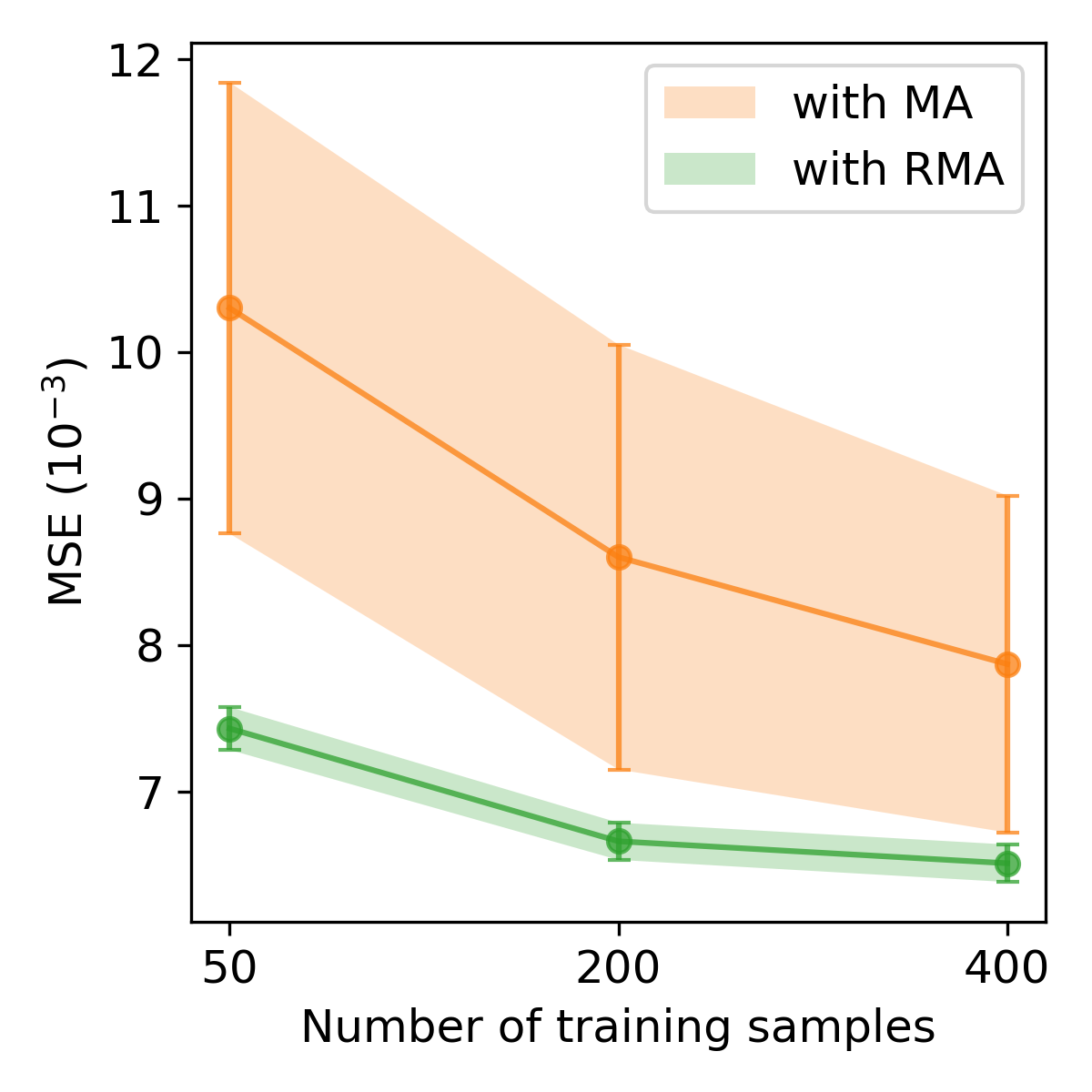}
    \caption{LPGD}
  \end{subfigure}
  \begin{subfigure}[t]{0.32\textwidth}
    \centering
    \includegraphics[width=0.95\linewidth]{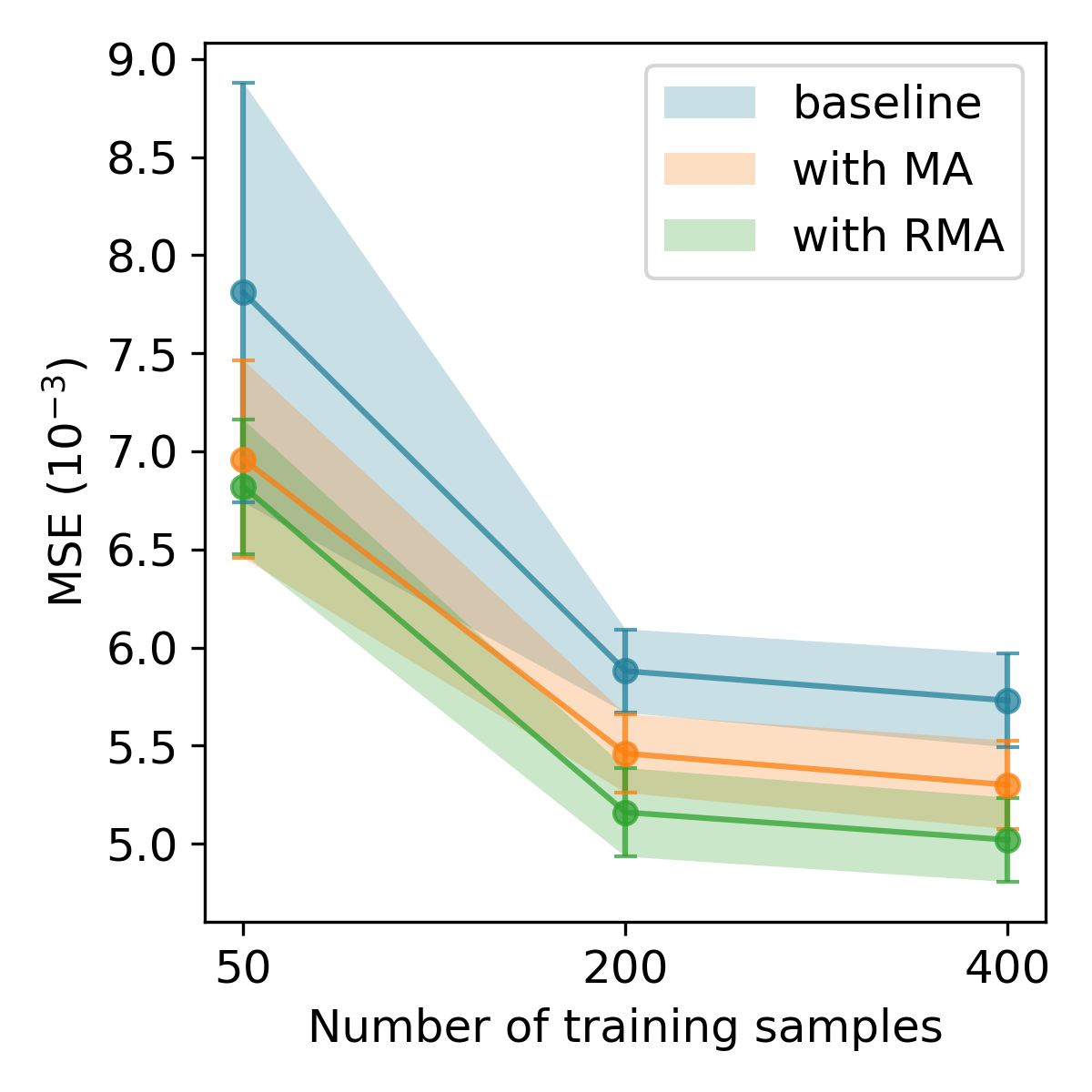}
    \caption{LPGDSW}
  \end{subfigure}
  \begin{subfigure}[t]{0.32\textwidth}
    \centering
    \includegraphics[width=0.95\linewidth]{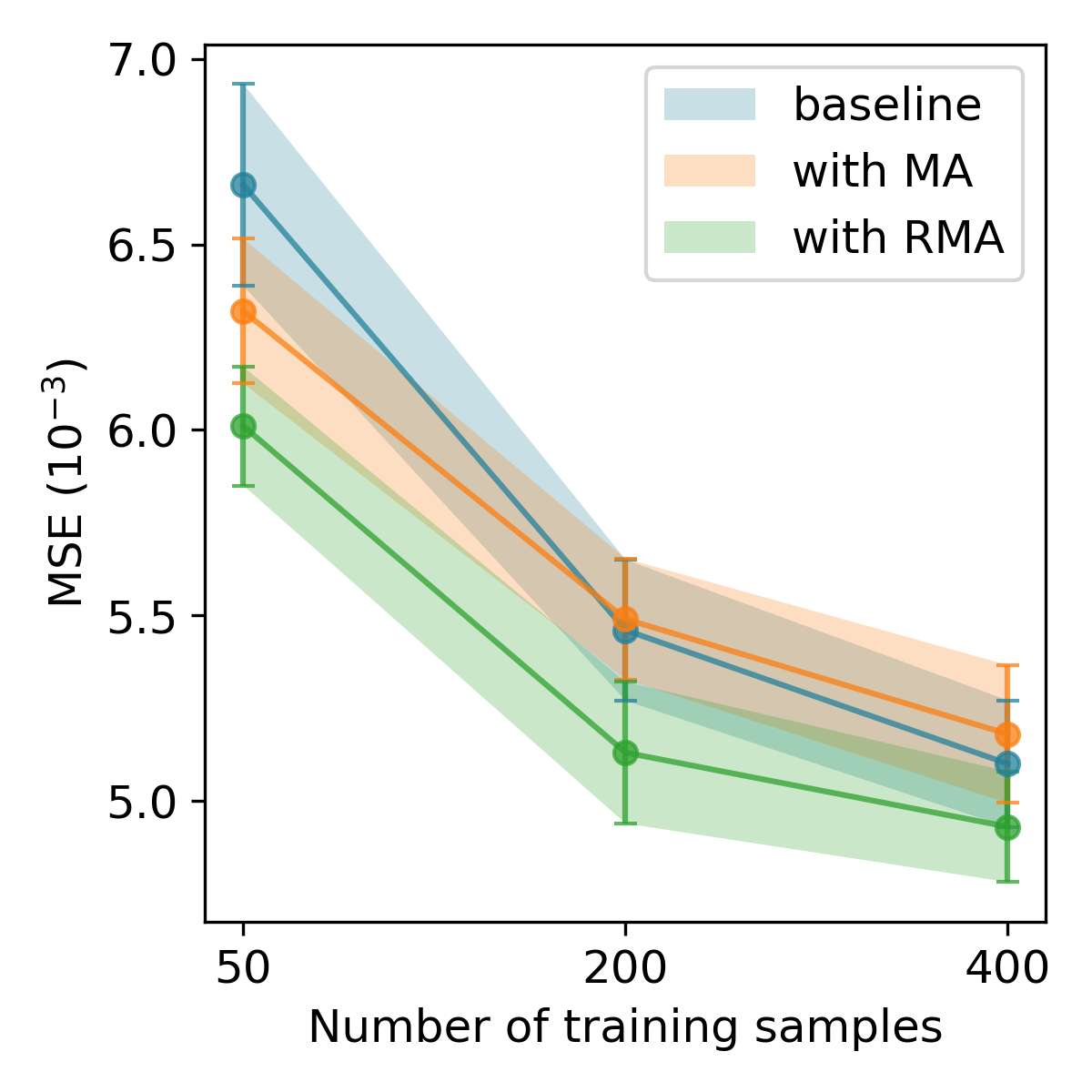}
    \caption{LPD}
\end{subfigure}
 \caption{ The MSE results plotted against the number of training data
 for Case 2. The solid line represents the average of 10 tests, and the shade around the solid line depicts the one standard deviation. }
  \label{fig:eit-4-differnt-training-number}
\end{figure}

\paragraph{Data size} 
Next we examine the performance of the DuNets methods with respect to the data size. 
For this purpose we visualize the results in Tables~\ref{tab:eit2targets} and~\ref{tab:eit-4-targets}
with Figs.~\ref{fig:eit-2-differnt-training-number} and~\ref{fig:eit-4-differnt-training-number}. 
One observes that MSE is decreasing with more training data used in all the methods with RMA, 
which is not the case for the baseline methods and those with MA. 
For example, in Case 1, the MSE results of LPGDSW become evidently higher when the data size changes from 200 to 400.  
These results demonstrate that the RMA module can increase the stability of the DuNet methods with respect the data size. 

\paragraph{Number of inclusions}
We now study how the methods perform with different numbers of inclusions by comparing the results in Case 1 (2 inclusions) and Case 2 (4 inclusions). 
We can see that the MSE results in Case 2 are generally higher than those in Case 1, indicating that more inclusions make the problems more challenging for all methods.
Nevertheless,  the RMA module considerably improves the performance of DuNets in both cases, an evidence that RMA is rather robust against the number of inclusions. 
In Case 1, the PDIPM-TV method yields a notably lower average MSE than that of LPGM-MA with 50 training datasets. Case 2 results show PDIPM-TV outperforming LPGM-MA, LPGDSW with 50 datasets, and LPGD-RMA with 200 datasets, thus affirming the benefits of sparsity regularization on model performance.

\begin{table}[!htb]
\centering
\caption{The average MSE values of the DuNets methods for Case 2, 
 with the associated standard deviations in parentheses. The average MSE for the GN approach is 12.6E-03 ($\pm$ 1.0E-03), and for the PDIPM-TV method, it is 7.67E-03 ($\pm$ 0.36E-03).
The best results are indicated in \textcolor{orange}{orange} color.
}\label{tab:eit-4-targets}
\begin{tabular}{llll}
\hline
data size &
  \multicolumn{1}{c}{50} &
  \multicolumn{1}{c}{200} &
  \multicolumn{1}{c}{400} \\  \hline 
LPGD &
  \multicolumn{1}{c}{---} &
  \multicolumn{1}{c}{---} &
  \multicolumn{1}{c}{---} \\ \specialrule{0em}{2pt}{2pt}
LPGD-MA &
  \begin{tabular}[c]{@{}c@{}}10.3E-03\\ ($\pm$ 1.54E-04)\end{tabular} &
  \begin{tabular}[c]{@{}c@{}}8.60E-03\\ ($\pm$ 1.45E-04)\end{tabular} &
  \begin{tabular}[c]{@{}c@{}}7.87E-03\\ ($\pm$ 1.15E-04)\end{tabular}  \\  \specialrule{0em}{2pt}{2pt}
LPGD-RMA &
  \begin{tabular}[c]{@{}c@{}}{\textcolor{orange}{7.87E-03}}\\ ($\pm$ 1.46E-04)\end{tabular} &
  \begin{tabular}[c]{@{}c@{}}{\textcolor{orange}{7.43E-03}}\\ ($\pm$ 1.27E-04)\end{tabular} &
  \begin{tabular}[c]{@{}c@{}}{\textcolor{orange}{6.66E-03}}\\ ($\pm$ 1.28E-04)\end{tabular} \\ \hline

LPGDSW &
  \begin{tabular}[c]{@{}c@{}}7.81E-03\\ ($\pm$ 1.07E-04)\end{tabular} &
  \begin{tabular}[c]{@{}c@{}}6.88E-03\\ ($\pm$ 2.13E-04)\end{tabular} &
  \begin{tabular}[c]{@{}c@{}}6.69E-03\\ ($\pm$ 2.39E-04)\end{tabular} \\ \specialrule{0em}{2pt}{2pt}
LPGDSW-MA &
  \begin{tabular}[c]{@{}c@{}}{6.96E-03}\\ ($\pm$ 5.04E-04)\end{tabular} &
  \begin{tabular}[c]{@{}c@{}}5.46E-03\\ ($\pm$ 2.01E-04)\end{tabular} &
  \begin{tabular}[c]{@{}c@{}}5.53E-03\\ ($\pm$ 2.26E-04)\end{tabular}  \\ \specialrule{0em}{2pt}{2pt}
LPGDSW-RMA &
  \begin{tabular}[c]{@{}c@{}}{\textcolor{orange}{6.82E-03}}\\ ($\pm$ 3.42E-04)\end{tabular} &
  \begin{tabular}[c]{@{}c@{}}{\textcolor{orange}{5.16E-03}}\\ ($\pm$ 2.26E-04)\end{tabular} &
  \begin{tabular}[c]{@{}c@{}}{\textcolor{orange}{5.02E-03}}\\ ($\pm$ 2.15E-04)\end{tabular}  \\ \hline \specialrule{0em}{2pt}{2pt}
  
LPD &
  \begin{tabular}[c]{@{}c@{}}6.66E-03\\ ($\pm$ 2.72E-04)\end{tabular} &
  \begin{tabular}[c]{@{}c@{}}5.46-03\\ ($\pm$ 1.90E-04)\end{tabular} &
  \begin{tabular}[c]{@{}c@{}}5.10E-03\\ ($\pm$ 1.71E-04)\end{tabular}\\ \specialrule{0em}{2pt}{2pt}
LPD-MA &
  \begin{tabular}[c]{@{}c@{}}6.32E-03\\ ($\pm$ 1.95E-04)\end{tabular} &
  \begin{tabular}[c]{@{}c@{}}5.49E-03\\ ($\pm$ 1.63E-04)\end{tabular} &
  \begin{tabular}[c]{@{}c@{}}5.18E-03\\ ($\pm$ 1.86E-04)\end{tabular}\\ \specialrule{0em}{2pt}{2pt}
LPD-RMA &
  \begin{tabular}[c]{@{}c@{}}{\textcolor{orange}{6.01E-03}}\\ ($\pm$ 1.60E-04)\end{tabular} &
  \begin{tabular}[c]{@{}c@{}}{\textcolor{orange}{5.13E-03}}\\ ($\pm$ 1.91E-04)\end{tabular} &
  \begin{tabular}[c]{@{}c@{}}{\textcolor{orange}{4.93E-03}}\\ ($\pm$ 1.49E-04)\end{tabular}  \\  \hline
\end{tabular}
\end{table}

\begin{figure}[!htb]
    \centering
    \includegraphics[width=0.6\linewidth]{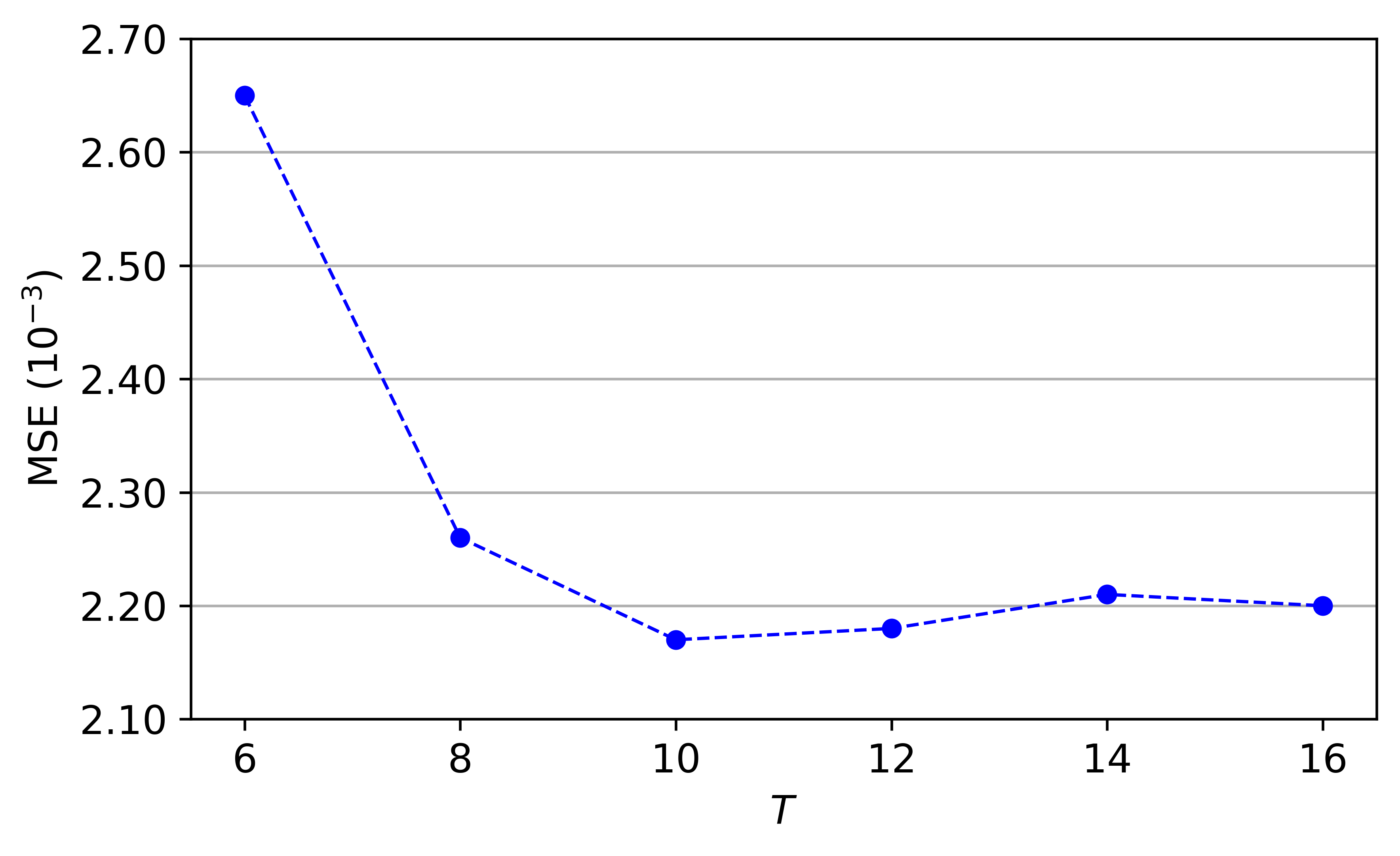}
    \caption{The MSE results plotted against the number of unrolled iterations.}
    \label{fig:mse_vs_T}
\end{figure}

\paragraph{Number of unrolled iterations}
Finally, we evaluate the impact of the unrolled iteration number $T$ on the performance of the LPD-RMA model. We train a set of LPD-RMA models with varying $T$ values ($T=6,8,\cdots,16$) using 200 training datasets in Case 1. 
Then we compute the average MSE value from ten independent runs over the 20 test datasets and present the results in Figure~\ref{fig:mse_vs_T}.  We can see from the figure that, when $T$ is less than 10, an increase in $T$ significantly enhances model performance in terms of the MSE value; however, when $T$ is greater than 10, this enhancement diminishes and the MSE value slightly rises. 
One possible reason for this phenomenon is that an increase in $T$ leads to a higher number of training parameters, making the model more prone to overfitting. Therefore, we should select an appropriate $T$ that represents a good balance between model performance and complexity.

\section{Conclusion}\label{sec:conclusion}
In this work, we propose a method for improving the performance of DuNets in nonlinear inverse problems. 
In particular, we apply RMA to two popular DuNets: LPGD and LPD respectively. 
We provide numerical experiments that can demonstrate the performance of the proposed method. 
We expect that the proposed method can be extended to other unrolling algorithms and applied to a wide range of nonlinear inverse problems. 

\section*{Acknowledgments}
The work is supported by the National Natural Science Foundation of China under Grant 12101614, and the Natural Science Foundation of Hunan Province, China, under Grant 2021JJ40715. We are grateful to the High Performance Computing Center of Central South University for assistance with the computations.

\appendix 
\section{The LSTM network}\label{sec:lstm}
\begin{figure}[!hb]
\centering
  \begin{subfigure}[t]{0.45\textwidth}
    \centering
    \includegraphics[width=0.95\linewidth]{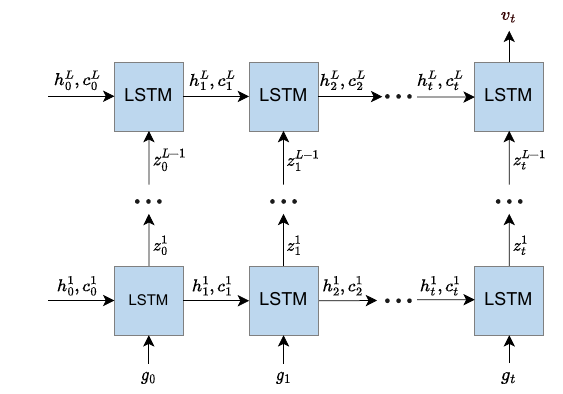}
    \caption{RMA structure}
  \end{subfigure}\label{fig:rma}
  \begin{subfigure}[t]{0.45\textwidth}
    \centering
    \includegraphics[width=0.95\linewidth]{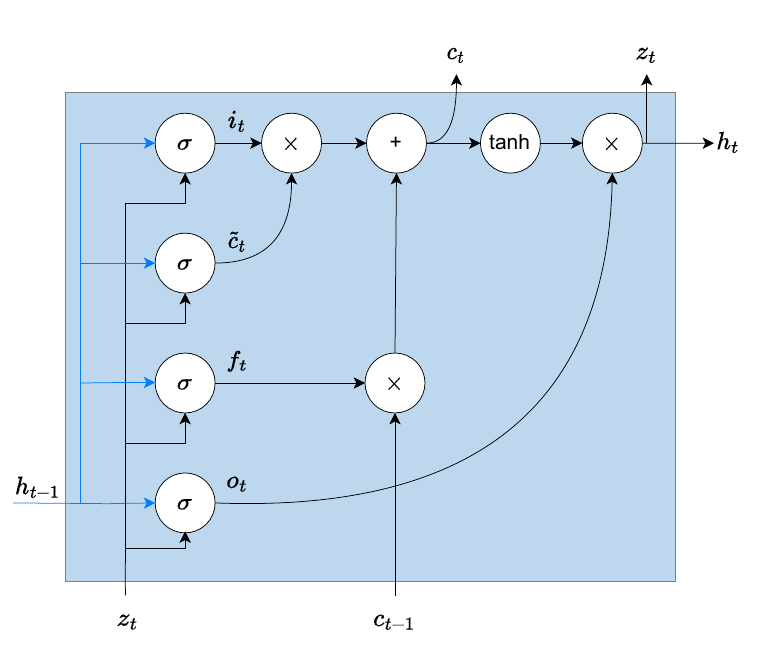}
    \caption{LSTM cell}
  \end{subfigure}\label{fig:lstm}
 \caption{
 (a) RMA structure uses a deep LSTM-RNN consisting of $L$ hidden layers. 
 (b) LSTM structure: cell gates are the input gate $\mathrm{i}_{t}$, forget gate $f_{t}$, output gate $o_{t}$, and a candidate cell state $\tilde{c}_{t}$. In practice, the current output $z_{t}$ is considered equal to the current hidden state $h_{t}$.
 }
\label{fig:rma-lstm}
\end{figure}

Here we will provide a detailed description of the LSTM network used in RMA. 
Recall that in RMA, at each time $t$, a network model $(v_{t}, h_{t},c_t) =\Xi_{\vartheta}\left(g_{t-1}, h_{t-1}, c_{t-1}\right)$ is used, and the structure of this network is specified as follows:
\begin{itemize}
\item The model $\Xi_{\vartheta}(\cdot)$ is a $L$-layer network with a LSTM-cell (denoted as $\mathrm{LSTM}^l(\cdot)$ for $l=1,...,L$) at each layer. 
\item Both the cell state $c_{t}$
and the hidden state $h_{t}$ have $L$ components: $c_t=(c^1_t\,,...,\,c^L_t)$ 
and $h_t=(h^1_t\,,...,\,h^L_t)$ with each component being a vector of a prescribed dimension,
and the initial states ${h}_{0}$ and $c_0$ are set to be zero.
\item For the $l$-th layer, the inputs of the LSTM cell are $g_{t}$, $c^1_{t-1}$ and $h^1_{t-1}$, 
and the output of it are  $h^1_{t}$ and an intermediate state $z^1_t$ that will be inputted into the next layer:
\begin{equation*}
 (z^{l}_t,h_{t}^{l},c_{t}^{l}) =\mathrm{LSTM}^l\left(z_t^{l-1}, h_{t-1}^{l}, c^l_{t-1}\right),
\end{equation*}
where $z_t^0$ is initialized as $g_{t}$ for the first layer $l=1$. In the final layer $l=L$, the output $h^L_{t}$ is set as the velocity $v_t$. 
\end{itemize}
The structure of the LSTM network is summarised in Fig.~\ref{fig:rma-lstm} (a). 

We now discuss the details of the LSTM cell that combines the input features $g_{t}$ at each time step and the inherited information from previous time steps.
In what follows we often omit the layer index $l$ when not causing ambiguity.
At each layer, the LSTM cell proceeds as follows.  
First LSTM  computes a candidate cell state $\tilde{c}_{t}$ by combining $h^l_{t-1}$ and $z^{l-1}_{t}$ (with $z^1_t=g_{t-1}$), as:
\begin{equation*}
\tilde{c}_{t}=\tanh \left(W_{hc} h^l_{t-1}+W_{gc} z^{l-1}_{t}+b_c\right),
\end{equation*}
and it then generates a forget gate $x_{t}$, an input gate $i_{t}$, and an output gate $o_{t}$ via the sigmoid function $\sigma(\cdot)$:
\begin{equation*}
\begin{aligned} f_{t} &=\sigma\left(W_{hx} h^l_{t-1}+W_{gx} z_t^{l-1}+b_x\right), 
\\ i_{t} &=\sigma\left(W_{hi} h^l_{t-1}+W_{gi} z_t^{l-1}+b_i\right), \\ o_{t} &=\sigma\left(W_{ho} h^l_{t-1}+W_{go} 
z_t^{l-1}+b_o\right). \end{aligned}
\end{equation*}
The forget gate is used to filter the information inherited from $c_{t-1}$, and the input gate is used to filter
the candidate cell state at $t$. Then we compute the cell state  $c_t$ via,
\[c_{t}=f_{t} \otimes c_{t-1}+i_{t} \otimes \tilde{c}_{t},\]
which serves as a memory reserving information from the previous iterations, and the hidden representation $h^l_t$ as, 
\begin{equation*}
\quad h_{t}=o_{t} \otimes \tanh \left(c_{t}\right),
\end{equation*}
where $\otimes$ denotes the element-wise product. 
Finally the output of the LSTM model $v_t$ at time $t$ is calculated as:
\begin{equation*}
v_{t} = W_{hg}h_t+b_g,
\end{equation*}
which is used to replace the gradient in the standard DuNets methods. 
This is a brief introduction to  LSTM  tailored for our own purposes, and readers who are interested in more details of the method may consult~\cite{hochreiter1997long,greff2016lstm}.

\section*{References}
\bibliographystyle{plain}
\bibliography{ref.bib}

\end{document}